\documentclass[10pt,journal,compsoc]{IEEEtran}
\IEEEoverridecommandlockouts 

\usepackage{booktabs}

  \usepackage[nocompress]{cite}
  
\usepackage{graphicx}
\ifCLASSINFOpdf
\else
\fi
\usepackage{bbm}
\usepackage{amsmath}
\usepackage{amsfonts,amssymb}
\usepackage{makecell}
\usepackage{colortbl} 
\usepackage{array}
\usepackage[most]{tcolorbox}
\usepackage{tikz,xcolor}
\usepackage{hyperref}
\usepackage{scalerel}
\usetikzlibrary{svg.path}

\hypersetup{hidelinks,
	colorlinks=true,
	allcolors=black,
	pdfstartview=Fit,
	breaklinks=true}
\definecolor{orcidlogocol}{HTML}{A6CE39}
\tikzset{
  orcidlogo/.pic={
    \fill[orcidlogocol] svg{M256,128c0,70.7-57.3,128-128,128C57.3,256,0,198.7,0,128C0,57.3,57.3,0,128,0C198.7,0,256,57.3,256,128z};
    \fill[white] svg{M86.3,186.2H70.9V79.1h15.4v48.4V186.2z}
                 svg{M108.9,79.1h41.6c39.6,0,57,28.3,57,53.6c0,27.5-21.5,53.6-56.8,53.6h-41.8V79.1z M124.3,172.4h24.5c34.9,0,42.9-26.5,42.9-39.7c0-21.5-13.7-39.7-43.7-39.7h-23.7V172.4z}
                 svg{M88.7,56.8c0,5.5-4.5,10.1-10.1,10.1c-5.6,0-10.1-4.6-10.1-10.1c0-5.6,4.5-10.1,10.1-10.1C84.2,46.7,88.7,51.3,88.7,56.8z};
  }}

\newcommand\orcidicon[1]{\href{https://orcid.org/#1}{\mbox{\scalerel*{
\begin{tikzpicture}[yscale=-1,transform shape]
\pic{orcidlogo};
\end{tikzpicture}
}{|}}}}
\begin{document}
%
\title{Uncertainty in Natural Language Processing: Sources, Quantification, and Applications}
%
%
%
%



\author{Mengting Hu, Zhen Zhang, Shiwan Zhao, Minlie Huang and Bingzhe Wu
\IEEEcompsocitemizethanks{
 
\IEEEcompsocthanksitem Mengting Hu and Zhen Zhang are with the College of Software, Nankai University. E-mail: mthu@nankai.edu.cn, zhangz@mail.nankai.edu.cn

\IEEEcompsocthanksitem Shiwan Zhao is an independent researcher. E-mail: zhaosw@gmail.com

\IEEEcompsocthanksitem Minlie Huang is with the Department of Computer Science, Tsinghua University. E-mail: aihuang@tsinghua.edu.cn

\IEEEcompsocthanksitem Bingzhe Wu is with Tencent AI Lab. E-mail: bingzhewu@tencent.com

}

\thanks{This work has been submitted to the IEEE for possible publication. Copyright may be transferred without notice, after which this version may no longer be accessible.}
}

\IEEEtitleabstractindextext{%
\begin{abstract}

As a main field of artificial intelligence, natural language processing (NLP) has achieved remarkable success via deep neural networks. Plenty of NLP tasks have been addressed in a unified manner, with various tasks being associated with each other through sharing the same paradigm. However, neural networks are black boxes and rely on probability computation. Making mistakes is inevitable. Therefore, estimating the reliability and trustworthiness (in other words, \emph{uncertainty}) of neural networks becomes a key research direction, which plays a crucial role in reducing models' risks and making better decisions. Therefore, in this survey, we provide a comprehensive review of uncertainty-relevant works in the NLP field. Considering the data and paradigms characteristics, we first categorize the sources of uncertainty in natural language into three types, including input, system, and output. Then, we systemically review uncertainty quantification approaches and the main applications. Finally, we discuss the challenges of uncertainty estimation in NLP and discuss potential future directions, taking into account recent trends in the field. Though there have been a few surveys about uncertainty estimation, our work is the first to review uncertainty from the NLP perspective.

\end{abstract}

}

\maketitle

\IEEEdisplaynontitleabstractindextext

%
\IEEEpeerreviewmaketitle

\IEEEraisesectionheading{\section{Introduction}\label{sec:introduction}}

\IEEEPARstart{N}{atural} Language Processing (NLP) is a multidisciplinary field that encompasses computer science, artificial intelligence, and linguistics. Its aim is to develop machines that understand natural language and allow humans to interact with it using natural language. Benefiting from the development of deep neural networks (DNNs), NLP technology has a wide range of applications, including sentiment analysis (SA), machine translation (MT), question-answering (QA) systems, etc. With the development of pre-trained language models (PLMs) \cite{sun2022paradigm}, plenty of NLP tasks can be tackled in a similar manner by sharing the same paradigm. That is to say, various tasks can be simply formulated into a classification, regression, generation, etc., problem. A unified trend is being observed in the development of NLP field, which enables the efficient utilization of PLMs and promotes knowledge transfer across tasks. 

\begin{figure}
    \centering
    \includegraphics[scale=0.5]{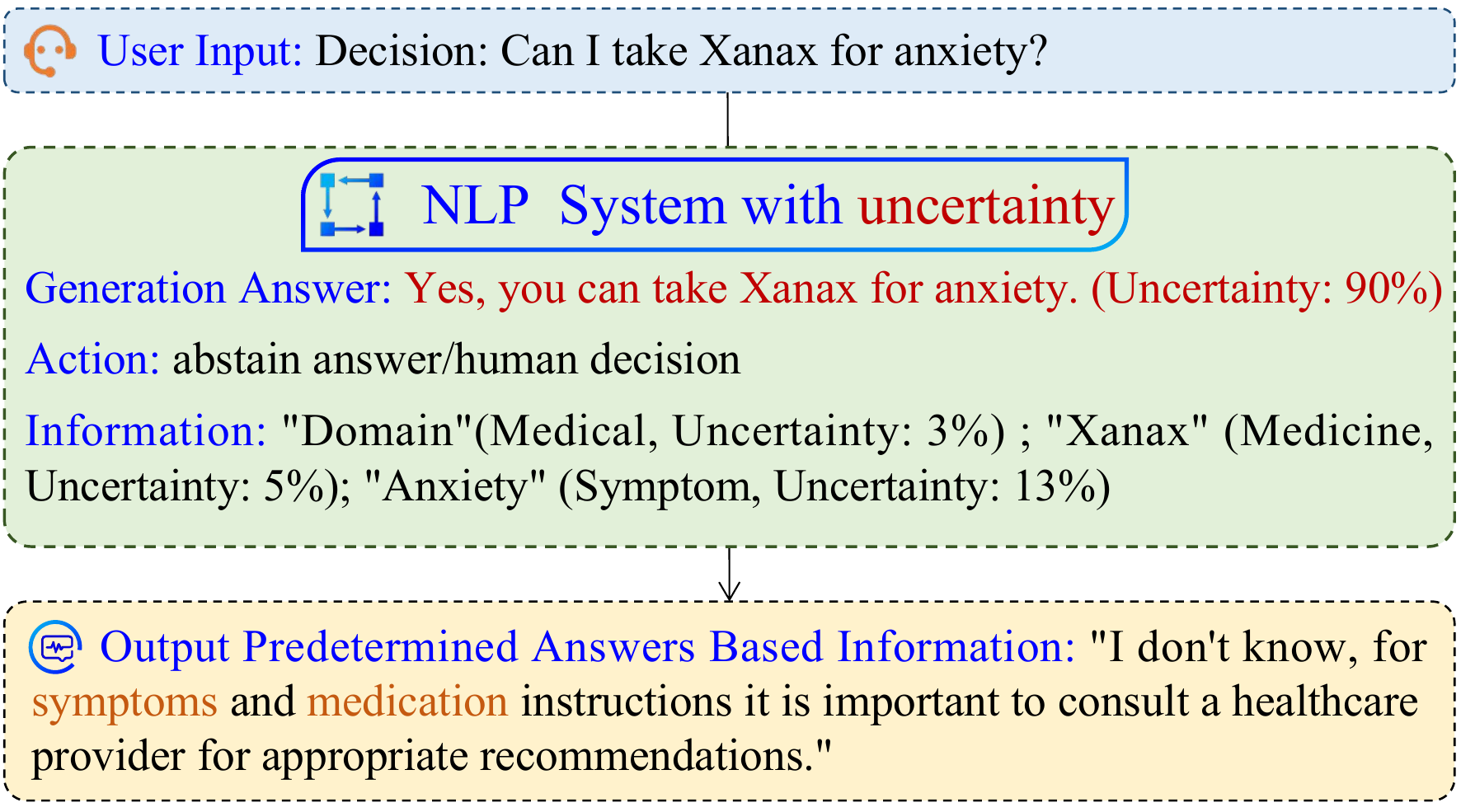}
    \caption{An illustration of NLP systems applied in the medical domain.}
    \label{fig:example_ue_pipl}
\end{figure}

Though NLP field has achieved great success, it relies heavily on neural networks. Such a technique is a black box and depends on probability computation. There must be probable to make mistakes, which is inevitable. Thus, estimating the uncertainty of neural networks becomes a crucial research direction, which measures the reliability and trustworthiness of models. Especially in safety-critical applications like autonomous systems or medical diagnosis, uncertainties in predictions can have severe consequences if not appropriately addressed. Here, we demonstrate the importance of uncertainty through an example in the medical domain. As depicted in Fig. \ref{fig:example_ue_pipl}, a user inquires about the potential use of the addictive drug \emph{``Xanax''} for anxiety, the system fails to differentiate the intensity of anxiety symptoms, resulting in a direct affirmation of its usability. This is clearly deemed inappropriate for patients, considering the grave side effects associated with the drug \emph{``Xanax''}. However, incorporating a probability of uncertainty would aid in determining whether to abstain from the response or defer to the expertise of professionals for a final decision. Some studies have shown that accurately quantifying uncertainty can help identify situations where the model is uncertain, thereby improving the reliability and interpretability of the output \cite{siddhantlipton2018deep,xiao2019uncerNLP,kuhn2023semanticeq}.

Considering the unified trends of NLP and the essentiality of uncertainty, a systematic review of uncertainty-relevant NLP and the corresponding solutions is still lacking, which we aim to fill in this survey. Initially, we arise a research question: what are the \textbf{sources} of uncertainty in the NLP field? Here, we naturally connect uncertainty with multiple stages of an NLP system and summarize the sources into \emph{input}, \emph{system}, and \emph{output}. More concretely, natural language input to NLP system is inherently ambiguous and context-dependent, making it difficult to achieve perfect performance and reliability in many NLP tasks \cite{ethayarajh2020classifier,kivimaki2022uncertainty,pei2022transformer}. Then, the system, such as the network initialization, architecture, and interior computation, etc., could introduce randomness, leading to uncertainty. Lastly, the uncertainties of outputs are due to complicating factors, such as a lack of broader knowledge, unreasonable model parameters, or specification of task output.

In light of the in-depth analysis of uncertainty sources, we further review the literature about the uncertainty \textbf{quantification}. Uncertainty estimation is ubiquitous in DNNs, and a common example is represented by confidence of the network output. Specifically, the Softmax scores obtained from these networks provide a direct means of estimating confidence values that are easily converted to uncertainty scores, e.g. subtracting the confidence from 1. However, the predictions are typically represented as point estimates \!\cite{malinin2018prior,kopetzki2021evaluating}, and mistakes are inevitable. Guo et al. \cite{guo2017calibration} point out that DNNs tend to exhibit excessive confidence, making their confidence scores inaccurate. This is problematic because under-/over-confident in the network can lead to wrong decisions and actions based on overconfident predictions. To address this challenge, researchers have developed various uncertainty estimation methods for DNNs. This survey categorizes uncertainty estimation methods into three groups based on different modeling approaches: (1) \textit{calibration confidence}-based methods, (2) \textit{sampling}-based methods, and (3) \textit{distribution}-based methods. These methods go beyond traditional confidence level measurements to quantify the uncertainty in model predictions, identify situations where NLP systems are uncertain about outputs, and offer insights into model behavior. 
Throughout the survey, we discuss the features and challenges associated with each of these estimation methods.


Recently, the \textbf{applications} of uncertainty estimation in NLP are diverse and increasing. In this survey, we mainly divide the applications into three categories, including (1) \textit{data filtering and action guidance}, which is to select the required data according to the uncertainty and take the next action. In typical active learning, filter the data through the uncertain estimation of the teacher model, and guide the student network to train; another example is the detection of out-of-distribution data. (2) \textit{Improve the performance and efficiency of the system} based on uncertainty, (3) \textit{Output quality assessment}, such as the quality assessment of machine translation and understanding whether the answer of the QA system is trustworthy.  

The challenges of uncertainty estimation in applications depend on the problem setting, including the paradigm of the NLP model and the type of task (e.g., classification, generation, regression, etc.). When observing the challenges faced by the three applications mentioned above, we considered different paradigms for classification in deterministic estimation methods in this survey. Recently, with the emergence of large pre-trained models such as T5 \cite{raffel2020T5} and the GPT family \cite{radford2018improving,radford2019language,brown2020language}, which can capture linguistic patterns and knowledge from massive text data. The capabilities of NLP models in various paradigms have greatly improved. However, the complexity of large models makes it difficult to understand the models themselves, which include \textit{high-dimensional spaces}, \textit{variable lengths of generated text}, and \textit{expressions of uncertainty }for interpretation of results. These will be described in detail in the survey \cite{xu2020understanding, kuhn2023semanticeq}. 

Overall, the main purpose of this survey paper is to systematically review the advances and challenges of uncertainty in the NLP field. To the best of our knowledge, our paper is the first to summarize this topic. Concretely, considering recent trends and paradigms in NLP, we provide a categorization of uncertainty sources according to the stages of NLP systems. Besides, a comprehensive review of the uncertainty estimation (a.k.a. quantification) technique in recent years is further provided. We also summarize the applications of uncertainty in NLP. Finally, we conclude the challenges based on the characteristics of text data and the NLP paradigms, highlighting future challenges and potential trends.



\subsection{Organization of The Survey}

\begin{figure*}
    \centering
    \includegraphics[scale=0.5]{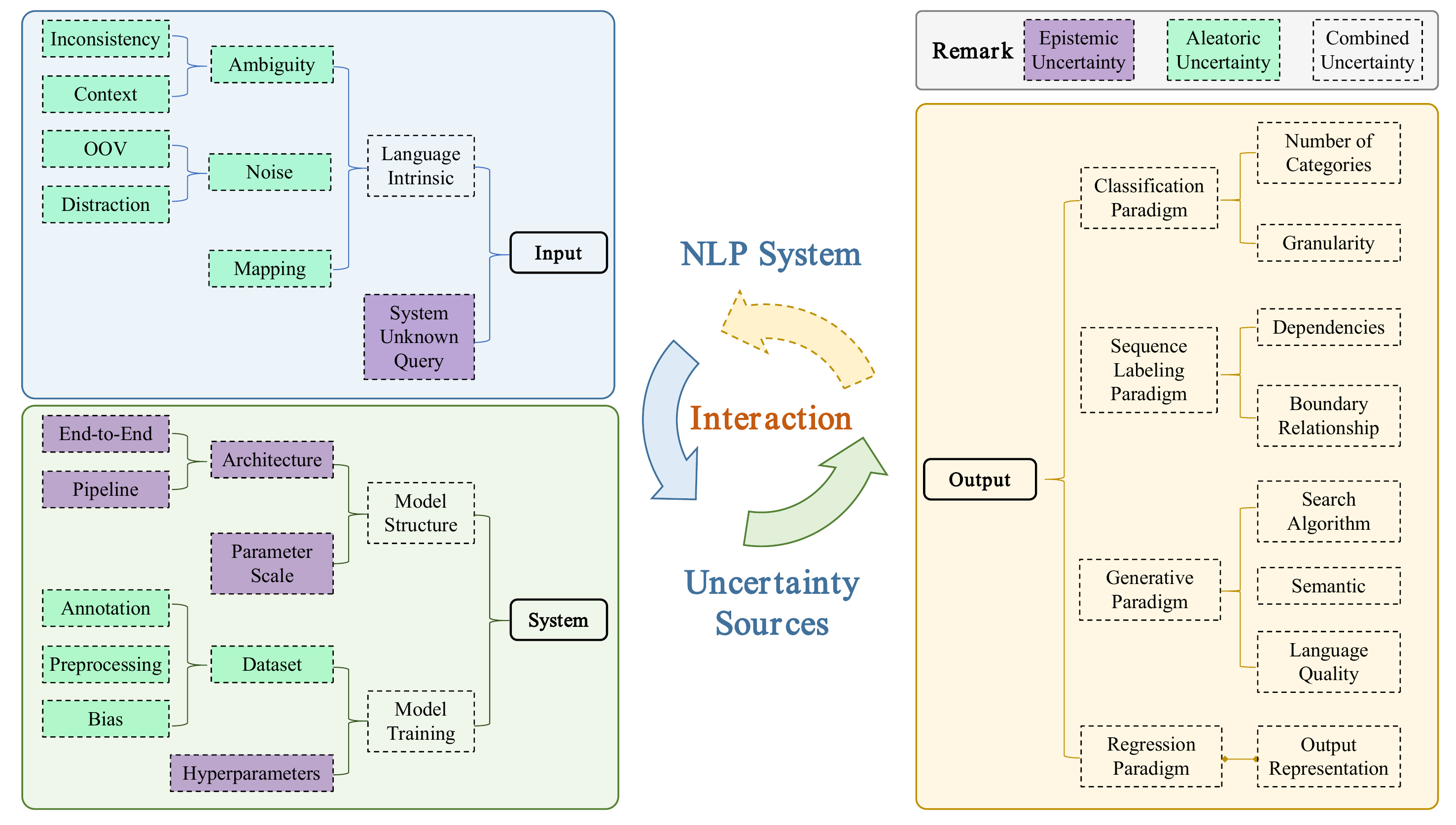}
    \caption{Illustration of sources of uncertainty. The figure includes the sources of uncertainty in the interaction of the NLP system. We start from the three processes of \textbf{Input}, \textbf{System}, and \textbf{Output} to analyze the possible causes of each uncertainty. It is worth noting that these three parts are interrelated. As the query passes through the NLP system, due to the combination of neural networks and different task specifications in complex ways, the type of uncertainty in the output prediction becomes complicated, including both the aleatoric and epistemic uncertainty, which are hard to decompose. Thus, we refer to it as \emph{combined uncertainty}.}
    \label{fig:uncer_sources}
\end{figure*}

The rest of the survey is organized as follows: In Sec \ref{Overview}, we introduce two kinds of uncertainty backgrounds in deep learning and illustrate the main sources of uncertainty in NLP systems in Fig. \ref{fig:uncer_sources}. In Sec \ref{UE_Technology}, we classify uncertainty estimation techniques based on their modeling approach. We also abstract the insights behind each category of technology and provide a comparison of different technologies. Following this, we present commonly used uncertainty evaluation metrics in NLP. Then, in Sec \ref{APP_UE_NLP}, we summarize the application directions of uncertainty in NLP under the three major categories and sort out the detailed literature of specific applications under each category. We then discuss the current challenges of uncertainty estimation and potential future research directions in Sec \ref{Challenges_Directions} and conclude this survey in Sec \ref{Conclusion}.

\section{Uncertainty Sources}\label{Overview}
In this section, we aim to analyze the uncertainty sources of NLP from the \emph{input}, \emph{system}, and \emph{output} stages, respectively. In advance, we review the theoretical background of uncertainty definition. 
\subsection{Theory Background of Uncertainty}
In the field of machine learning, uncertainty is commonly divided into aleatoric uncertainty and epistemic uncertainty (\!\cite{xiao2019uncerNLP,gawlikowski2021dnnUncerSurvey,hullermeier2021aleatoric,kivimaki2022uncertainty}). In general, these two types of uncertainty can be summarized as follows:

\begin{itemize}
    \item \textbf{Aleatoric Uncertainty} \; It is also known as data uncertainty, which refers to the uncertainty inherent in data due to its randomness or noise. This type of uncertainty is irreducible, meaning it cannot be eliminated through model improvements or tuning. It can arise from a variety of sources, such as noisy observations, overlapping classes, ground truth errors, inherent randomness, or other factors that are not entirely predictable.
    \item \textbf{Epistemic Uncertainty} \; It is also known as model uncertainty, which is reducible uncertainty that arises from a lack of knowledge or understanding about the model itself. This can include uncertainty about the model's structure choice, and parameters, which can be reduced by increasing the amount and quality of training data. Epistemic uncertainty can also result from out-of-distribution examples, such as new languages or domains, and may be caused by model structure errors or training procedure errors.
\end{itemize}

It is worth noting that there is no clear distinction between epistemic uncertainty and aleatoric uncertainty, and these types of uncertainty can even be mutual \cite{hullermeier2021aleatoric}. In particular, it may not always be clear which type of uncertainty is dominant in a given NLP system, and the two types of uncertainty can interact in complex ways, thus making combinations complex and difficult to decouple. Nonetheless, it is important to understand the source and classification of uncertainty, e.g., in MT, decoupled uncertainty can be used to infer whether predictions originate from noisy and ambiguous references, or out-of-distribution examples or noisy annotations \cite{zerva2022disentangling}.

To better understand the sources of uncertainty in NLP systems, it is necessary to incorporate NLP usage scenarios. To this end, we consider human-machine interactions in NLP tasks to mine factors that may contribute to uncertainty. Ultimately, we have determined three main sources of uncertainty: (1) \textbf{Input} text entered by a user, (2) \textbf{System} composed of models, and (3) \textbf{Output} returned to the user. In the following sub-parts, we will discuss these three sources. Meanwhile, in Fig. \ref{fig:uncer_sources}, we provide a detailed source overview. It's worth emphasizing that three sources tend to be interconnected, and addressing one may help to mitigate uncertainty in the others. 

\subsection{Uncertainty Source from Input}

During the usage of an NLP system, a user would feed input to it, expecting to obtain processed results or interactive feedback. Due to the language's intrinsic ambiguity and unknown query, the input itself contains uncertainty. Formally, assuming that the model parameters $\mathbf{W}$ have been trained on a dataset $D = \{(\mathbf{x}_{1}, y_{i})\}^{N}_{i=1}$, and there exists a correct and specific hypothesis space $\mathcal{H}$, if an input text falls within the hypothesis space $\mathcal{H}$, then the input adheres to the data distribution. However, if an input contains inherent uncertainty or noise of the language, it may fall on the boundary of the hypothesis space or even outsides \cite{delaforge2022ebbe}. In such cases, the system is prone to making incorrect judgments. Consequently, it is necessary to capture and quantify this resulting uncertainty. As depicted in the upper-left of Fig. \ref{fig:uncer_sources}, we distinguish common sources of uncertainty in inputs from two perspectives, including the language intrinsic and unknown query, respectively.



\subsubsection{Language Intrinsic}

Language inherently contains \textbf{ambiguity} that leads to uncertainty \cite{dragos2013ontological}. That is to say, ambiguity is a natural feature of language that arises from inconsistencies and contextual factors in the text \cite{blodgett2020Inconsistencies}. For instance, the word \emph{``apple''} has various meanings when appearing in different contexts, i.e. \emph{``I like to eat apple''} and \emph{``Steven Jobs built up Apple``}. Inconsistencies can take the form of incomplete sentences or expressions \cite{blodgett2020Inconsistencies}, while contextual semantics encompass the completeness and accuracy of information expression, as well as social and cultural aspects and differences between the unique context of text production (including features of space, time, authorship, etc.) and multiple interpretation contexts \cite{Venuti2008Translator,dragos2013ontological,ott2018MTuncer}. 

Natural language \textbf{noise} contains out-of-vocabulary (OOV) and distraction. OOV indicates spelling errors or emerging words that are unseen tokens for a running NLP system. These would result in unsureness. Besides, distracting refers to the presence of some words in a sentence that are not related to the meaning of the sentence. This kind of noise is ubiquitous in informal texts \cite{levy-2008-noisy,liu-etal-2018-word_noise,kadavath2022largeModelKnow}. For example, Liu et al. \cite{liu-etal-2018-word_noise} in entity recognition find that a part of the original sentence retains enough words to express the relationship, and the rest has many irrelevant words that can be regarded as noise that may hinder the performance of the extractor. Kadavath et al. \cite{kadavath2022largeModelKnow} set different prompt templates, such as helpful, incorrect, and distracting hints, and find that distracting hints in natural language generation (NLG) also lead to deviations in the generated answers.

As mentioned above, natural language is inherently complex, ambiguous, and context-dependent. It can be expressed in different ways to convey the same ideas or concepts. Therefore, the \textbf{mapping} relationship in NLP is also a source of uncertainty \cite{ott2018MTuncer}. This complexity can result in multiple input-output mapping scenarios, such as one-to-one and many-to-one mapping. In such cases, the model must select an output from several possible ones, which can be challenging and uncertain. For instance, in MT, the same source sentence can have multiple semantic equivalent translations, resulting in multiple nature of machine translation learning tasks \cite{Venuti2008Translator, ott2018MTuncer, wei2020uncertainty}. Similarly, Text-to-SQL needs to deal with many-to-one cases, indicating that multiple input texts correspond to the same SQL query \cite{qin2022sun}. This is also caused by the uncertainty of natural language. 

\subsubsection{System Unknown Query}

In the application of the actual world, systems usually encounter an unknown query differently distributed from training data \cite{gawlikowski2021dnnUncerSurvey, arora-etal-2021-types}. Systems tend to produce unreliable or even catastrophic predictions, thereby harming the trust of users. The uncertainty of the field is derived from the model that the model cannot explain the out-of-distribution (OOD) sample due to the lack of external knowledge. The source of this uncertainty is the input data extracted with the unknown subspace. Although DNNs can extract the knowledge in the domain from the domain migration sample, they cannot extract the knowledge samples in the domain from the external sample \cite{gawlikowski2021dnnUncerSurvey}. For example, QA systems deployed in search engines or individual assistants require elegantly processing OOD input, because users usually propose a problem that is not within the scope of system training distribution \cite{kadavath2022largeModelKnow,kamath-etal-2020-selective}. Another example is to make entailment judgments for breaking news articles in search engines \cite{carlebach-etal-2020-news}. If an input is not acceptable, the model needs to acknowledge its uncertainty and turn to humans or better (but more costly) models \cite{xinetal2021select2art}.


\subsection{Uncertainty Source from System}
The sources of uncertainty induced by NLP systems mainly originate from model blocks, which are usually introduced in the design and training of neural networks. The design of DNNs involves explicit modeling of the network architecture and its stochastic training process. The assumptions made about the problem structure, based on the network design and training, are referred to as inductive biases \cite{battaglia2018relational}.

\subsubsection{Model Structure}
Since neural networks are manually designed, their structures are also a source of uncertainty. In other words, changing the structure of networks might alter decision-making, bringing randomness to model training and inference stages. Here we discuss from the views of \emph{architecture} and \emph{parameter scale}.

In an end-to-end model, many factors are relevant to model architecture, such as the dimension of hidden states, and interior neuron connections, directly affect the performance of the network and thus introduce uncertainty \cite{lakshminarayanan2017simple, battaglia2018relational,gawlikowski2021dnnUncerSurvey} When studying complex AI systems, it needs to take into account the predictions of several independent models in downstream tasks \cite{amini2020deep} due to the effect of \textit{pipeline propagation}. The cost and risk of any decision-making process that produces different errors need to be balanced. This creates uncertainty when confidence in one module's predictions is allowed to have an impact on the next module. For this reason, if the uncertainty and confidence scores can be reliably interpreted as probabilities, the rules of probabilistic calculus can be applied, allowing one system to abort the decision if its predictions are not confident enough \cite{liu-etal-2020-fastbert,guietal2020uncertainty,schuster2022confident}. This is a useful property in many situations. For example, in the prediction of an information extraction system \cite{kivimaki2022uncertainty}, there are locations where the information extraction system extracts useful information and uses confidence to check whether the extracted information is meaningful. After this, another model makes a final prediction of the extracted value of the field. 

The parameter scale is a critical factor that contributes to uncertainty in a fixed model structure \cite{fedus2022switch}. The size of a model can have a significant impact on the level of uncertainty. In particular, larger models tend to possess a greater number of parameters and increased complexity. This enables them to better capture the intricacies of the training data, leading to reduced error on the training set. However, the larger parameter count also renders these models more susceptible to overfitting \cite{guo2017calibration}, potentially causing poor performance when presented with unseen data. On the other hand, smaller models have fewer parameters and lower complexity, making them more prone to underfitting the training data. Consequently, these models may exhibit higher error rates on the training set but have the potential to generalize better on previously unseen data, thereby potentially reducing uncertainty. It is important to recognize that the relationship between model size and uncertainty is multifaceted, and a comprehensive evaluation should consider various factors such as model complexity, dataset size and characteristics, and the regularization techniques employed during training.



\subsubsection{Model Training} 
During the model training, we distinguish uncertainty sources into the dataset and hyperparameter views, respectively. Initially, to learn an intelligent system, high-quality training data is essential. Yet due to the different language abilities of annotators, texts in the dataset pose various challenges for them \cite{beck2020representation}. As a result, this directly affects the correlation between samples and corresponding labels. Uncertainty arises whenever there are multiple possible interpretations of the data, but the knowledge or information to definitively choose one of them is not available \cite{bonneau2014overview}. Then, uncertainty may be part of the pre-processing stage. For example, NLP tools used for pre-processing can introduce uncertainty \cite{John2017UncertaintyIV}. More concretely, Part-of-speech taggers are often trained on data from historical periods. Yet, recent data is often not available. Making results for recent data is possibly unreliable, leading to uncertainty, and this uncertainty is usually irreducible. Furthermore, Bender et al. \cite{bender-friedman-2018-data} discuss ethical issues raised by PLMs, including issues of bias and fairness. They argue that language models encode biases and assumptions in the training data, which perpetuate social inequalities. This highlights sources of uncertainty related to the ethical implications of PLMs, and the need to consider them carefully for transparency.

The setting of hyperparameters, such as batch size, learning rate, and epoch number, is also random, resulting in different local optimal solutions and different final models \cite{gawlikowski2021dnnUncerSurvey}. Dodge et al. \cite{dodge2020fine} study the fine-tuning process of PLMs and find that hyperparameters and tuning such as weight initializations, data orders, and early stopping are critical to achieving high performance on downstream tasks. This highlights that under the large model, fine-tuning the model requires more strategies and hyperparameter requirements and adjustments, which will also introduce new uncertainty in the model framework.


\subsection{Uncertainty Source from Output}


Since making mistakes is inevitable for an NLP system, its output is not fully trusted, i.e. presenting uncertainty. Here we summarize the uncertainty of the output according to the paradigms and the nature of the task, namely \emph{classification}, \emph{sequence labeling}, \emph{generation}, and \emph{regression} paradigms. It is worth noting that although the task outputs are quite different, their sources of uncertainty are often combined and complex. That is to say, uncertainties sourced from the input text and model itself will affect the output. Therefore, in the right part of Fig. \ref{fig:uncer_sources}, we do not specifically define whether the uncertainty of output is epistemic or aleatoric uncertainty.

\subsubsection{Classification Paradigm}
The output of the classification paradigm is usually connected with the number and granularity of categories. Firstly, as the number of categories increases, the classification task becomes more difficult. There may be more errors or misclassification space, which enlarges uncertainty. This is because models may have difficulty distinguishing between similar categories, or be more error-prone due to the inherent ambiguity of the classification task. 
Secondly, the classifying granularity (e.g., classifying at words or sentences level, defining hierarchical or multi-label categories) may introduce uncertainty. More concretely, word-level classification may have variability. For example, a word like \emph{``Washington''} could be classified as a person or place depending on the context, and this classification itself may be vague or indeterminate. Classification at a coarser level (e.g. classifying entire sentences or documents) depends on more factors (e.g. contribution of each token, contextual semantics, length of sentence/document, etc.). 

\subsubsection{Sequence Labeling Paradigm}
Sequence labeling aims to assign a tagger to each token in the given input text. It needs to consider contexts to determine the dependencies of label sequences. Specifically, for sequence dependencies, the labels assigned to elements can depend on the labels assigned to adjacent elements in the sequence. This can lead to non-determinism because (1) due to the inherent non-determinism of natural language itself, the label assigned to an element may be ambiguous, which needs to be determined according to the context and neighboring elements in the sequence \cite{OpenNER,pandey2022multilinguals}. (2) The labels assigned to adjacent elements in the sequence should be consistent and follow a certain pattern. However, this can be challenging as the model may struggle to capture and model complex dependencies and relationships between adjacent elements \cite{panchendrarajan2018crfner}.

In addition, the boundary, e.g. the start and end positions of entities in named entity recognition (NER) tasks, is also crucial. Correct entity boundaries are effective in mitigating error propagation in entities that are linked to knowledge bases \cite{Li2022NERsurvey}.  Some studies consider entity boundary detection as a subtask in NER \cite{zhai2017neural,partalas-etal-2016-learning}. In summary, analyzing the prediction uncertainty in sequence labeling can be roughly divided into boundary uncertainty and boundary entity classification uncertainty. Both of these are inseparable from boundary detection and are dependent on the sequence. This analysis increases the rationality and reliability of the output results.

\subsubsection{Generation Paradigm}

A more challenging output is the generation paradigm, where the uncertainty of the prediction is affected by uncontrollable factors. The output space of generating natural language has $\mathcal{O}(|\mathcal{T}|^{N}$) dimensions. For example, MT models have hundreds of millions of parameters, and the search space is exponentially large, which poses great challenges for \textbf{search algorithm}. We usually observe only one reference for a given source sentence, leading to uncertainty in the final output. At present, tools and strategies can be borrowed and combined from machine learning and statistics \cite{levygoldberg2014linguistic,kuleshov2015calibrated,guo2017calibration,ott2018MTuncer}. Beam search is an effective search strategy, but some external uncertainty (such as the quality of training data) will lead to large beam performance degradation \cite{ott2018MTuncer}. 

One of the goals of NLG is to ensure that the generated text conveys the intended \textbf{semantic} content of the sentence. However, in decision-making tasks that rely on NLG, e.g. QA system \cite{kuhn2023semanticeq}, it is crucial to ensure invariance to the output space, but this is often not explicitly specified in the model. The meaning of the generated text is especially important in terms of its beliefs. While a system may be reliable even when generating output using many different expressions, it may be less reliable when answering questions with inconsistent meanings. 

Another key purpose of NLG is to ensure the \textbf{language quality}. which is typically evaluated based on several factors, including adequacy, fluency, readability, and variation \cite{stent2005evaluating}. A recent example of an effective strategy is reinforcement learning from human feedback (RLHF) \cite{bai2022training}. But it has been found to be wrongly calibrated \cite{kadavath2022largeModelKnow}. This is not surprising because RL fine-tuning tends to collapse language model predictions to the behavior that yields the highest reward, which raises the inherent problem of whether the model's output is biased or unfair. Furthermore, the length of the input text can vary widely \cite{murray2018correcting,malininuncertainty}, leading to uncertainty in the output. This particularly exists for longer sequences, where the joint likelihood decreases due to the conditional independence of labeling probabilities \cite{kuhn2023semanticeq}. 
Rare words also contribute to this problem because they have small probabilities and therefore occur less frequently in the sequence \cite{mikolov2013distributed, ott2018MTuncer}. 

\subsubsection{Regression Paradigm}

The uncertainty source in the classification output can be easily explained. A discrete probability distribution over the classifier's output is naturally provided through a softmax layer. However, in the regression paradigm, the output is a single numerical value in continuous target spaces \cite{wang-etal-2022-uncertainty}. The representation of the output becomes important, specifically the range of the output. For example, FLORES \cite{guzman-etal-2019-flores} divides 0-100 points into ten intervals, each representing different translation quality. In semantic text similarity assessment \cite{corley-mihalcea-2005-measuring}, the objective is to predict a similarity score for a sentence pair $({S}_1, {S}_2)$ within the range $[0, 5]$. A score of 0 indicates complete dissimilarity, while a score of 5 signifies the sentences are equal in meaning. Calibrated regression \cite{kuleshov2018accurate} extends the calibration method of classification to regression. It applies the obtained algorithm to the Bayesian deep learning model and calibrates the confidence score in the $[0, 1]$ interval.

\begin{figure}
    \centering
    \includegraphics[scale=0.5]{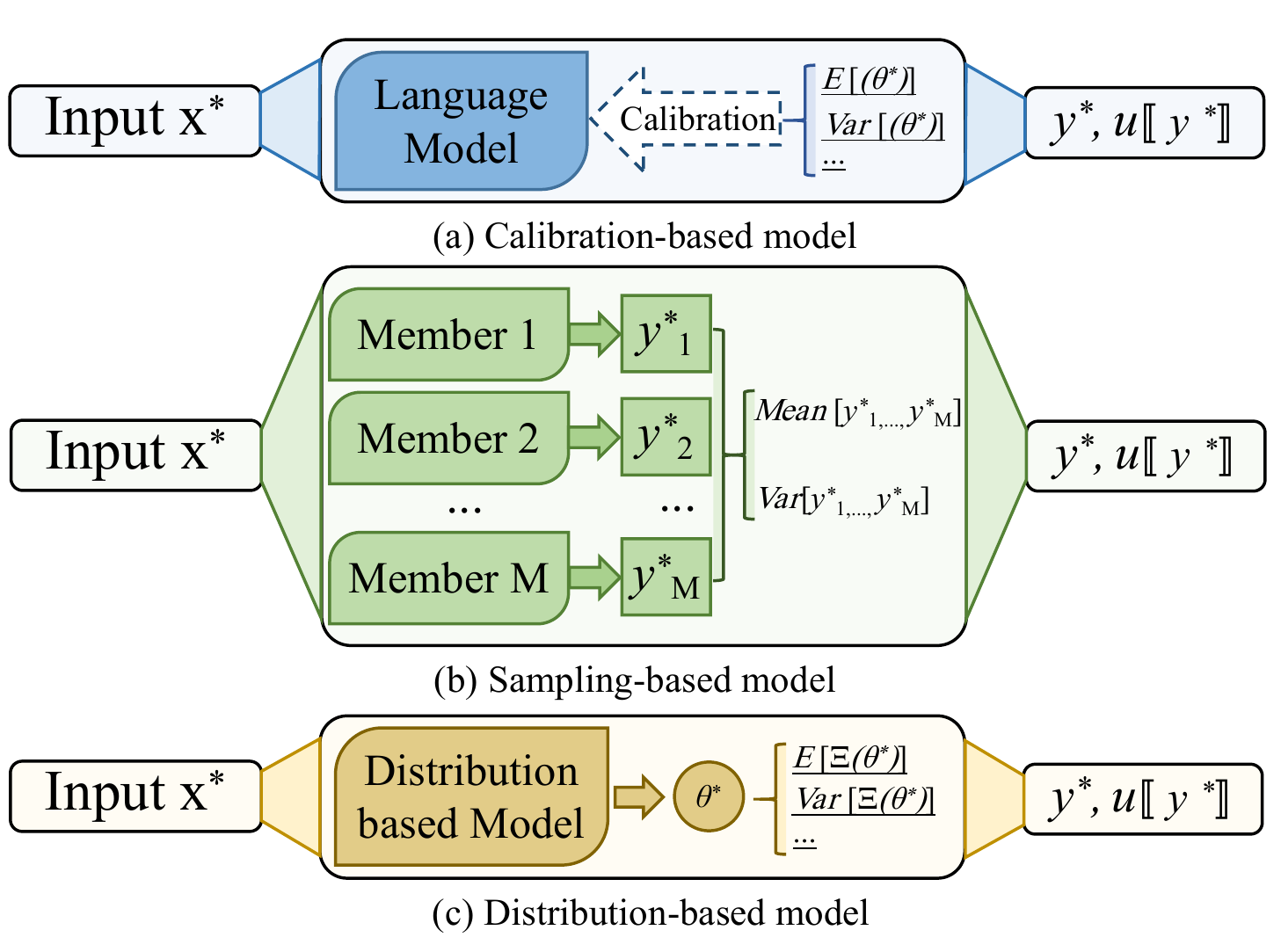}
    \caption{Rationale visualization of three types of uncertainty modeling. For a given input sample $\mathbf{x}$, each method provides a prediction y whose uncertainty is quantified as ${u}$. (a) Calibration-based uncertainty representation, (b) Sampling-based uncertainty estimation method, (c) Distribution-based uncertainty estimation method. $\Xi$ denotes a specific distribution. The mean $E$ and variance $Var$ are only used to keep the visualization simple, there are other quantifications in practice.}
    \label{fig:uncer_model}
\end{figure}

\section{Uncertainty Estimation}\label{UE_Technology}

Summarizing the sources of uncertainty in NLP systems can help reduce uncertainty, however, accurately distinguishing uncertainty types is challenging \cite{gawlikowski2021dnnUncerSurvey}. In this section, We first introduce the background of uncertainty estimation techniques, Then, As shown in Fig. \ref{fig:uncer_model}, according to the modeling characteristics, we propose a taxonomy of uncertainty estimation in NLP, and the specificity of linguistic uncertainty modeling.


\subsection{Uncertainty Estimation Background}

Modern neural networks are parameterized by a set of model weights $\boldsymbol{W}$, providing a formalization of uncertainty in the BNN case, and these are sufficient
statistics $\boldsymbol{\omega}=({{\mathbf{W}}_i})^{L}_{i}$. For a given dataset $\mathcal{D}=\{x_{i}, y_{i}\}^{N}_{i=1}$ with distribution ${{p}}(\mathbf{x},\mathbf{y})$, for a classification model ${{P}}({y}^{\ast}=\omega_{c}|{x}^{\ast},\mathcal{D})$ trained on $\mathcal{D}$, the distribution on prediction ${y}^{\ast}$ is described as:
\begin{equation}
    {{P}}({y}^{\ast}=\omega_{c}|\mathbf{x}^{\ast},\mathcal{D}) = \int \underbrace{{P}({y}^{\ast}=\omega_{c}|\mathbf{x}^{\ast},\boldsymbol{\theta}
    )}_{\text{(a) Data}}\,\underbrace{{p}(\boldsymbol{\theta}|\mathcal{D})}_{\text{(b) Model}}\,d\boldsymbol{\theta},
    \label{posterior_bnn}
\end{equation}
where the aleatoric uncertainty is formalized as the posterior probability ${{P}}({y}^{\ast}=\omega_{c}|\mathbf{x}^{\ast},\boldsymbol{\theta})$ of over class labels for a given parameter, while ${{p}}(\boldsymbol{\theta}|\mathcal{D})$ is used as the posterior distribution of the model parameters, describing the uncertainty of the model parameters, given a dataset $\mathcal{D}$.

However, ${{p}}(\boldsymbol{\theta}|\mathcal{D})$ is intractable using Bayes' rule posterior distributions, and it is necessary to use variational inference methods to achieve ${{p}}(\mathbf{\boldsymbol{\theta}}|\mathcal{D}) \approx q(\boldsymbol{\theta})$ \cite{blundell2015weight,kingma2015variational,louizos2016structured}. In the uncertainty estimation method, three ways are mainly considered, which will be placed in the next subsequent section. 

\subsection{Uncertainty Estimation Methods}
\begin{figure}
    \centering
    \includegraphics[scale=0.45]{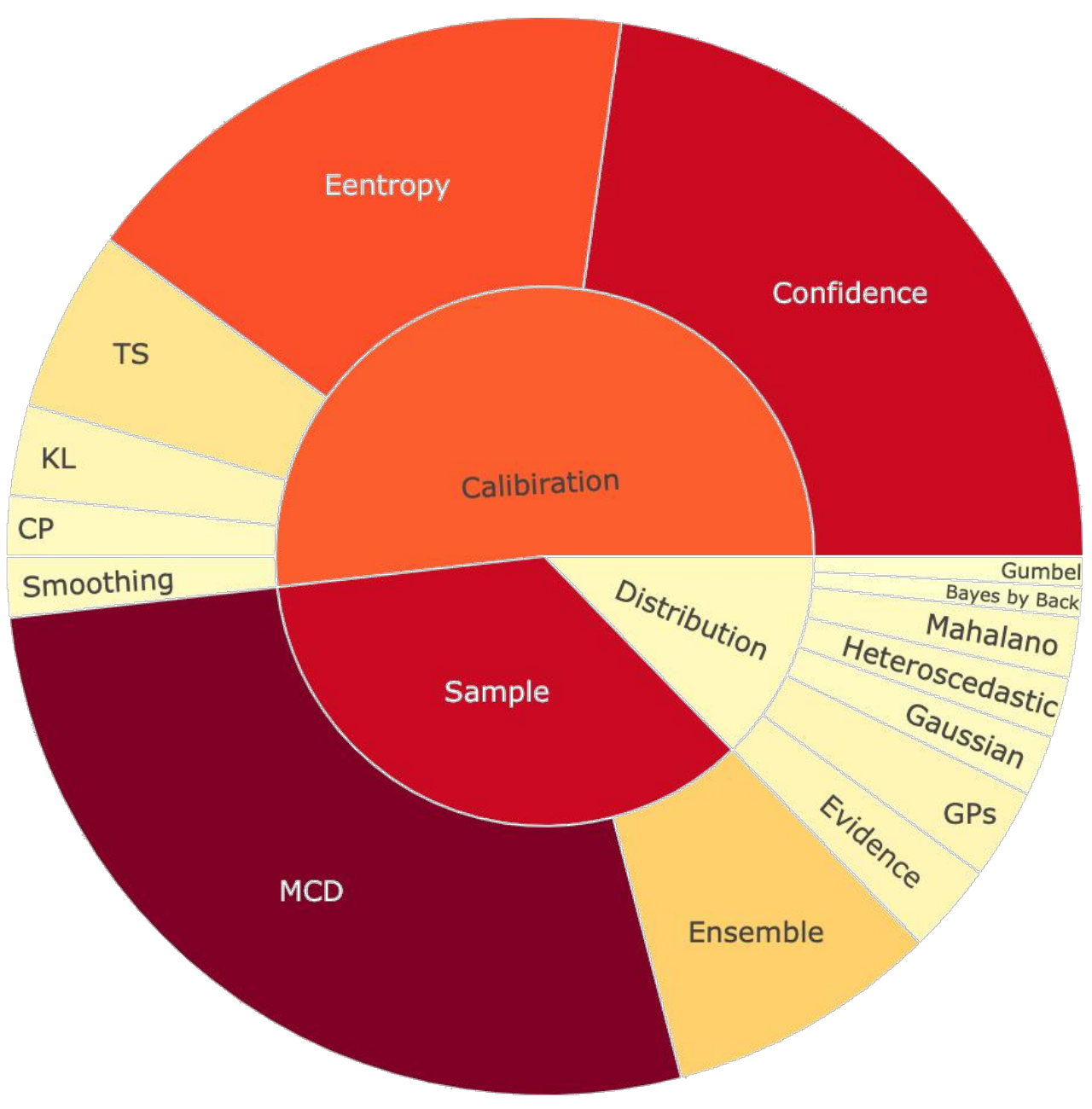}
    \caption{An overview of the taxonomy of uncertainty estimation techniques. The inner circle represents uncertainty modeling methods, the outer circle represents actual uncertainty methods, and some methods are represented by abbreviations.}
    \label{fig:statistics}
\end{figure}

After reviewing the uncertainty estimation methods commonly used in recent NLP tasks, we generally divide them into three types, namely, calibration confidence-based methods, sampling-based methods, and distribution-based methods. We provide the usage and hierarchical classification of these methods in Fig. \ref{fig:statistics}, and below we summarize from each of these three types.

\subsubsection{Calibration Confidence-based Methods} 
\begin{figure}
    \centering
    \includegraphics[scale=0.35]{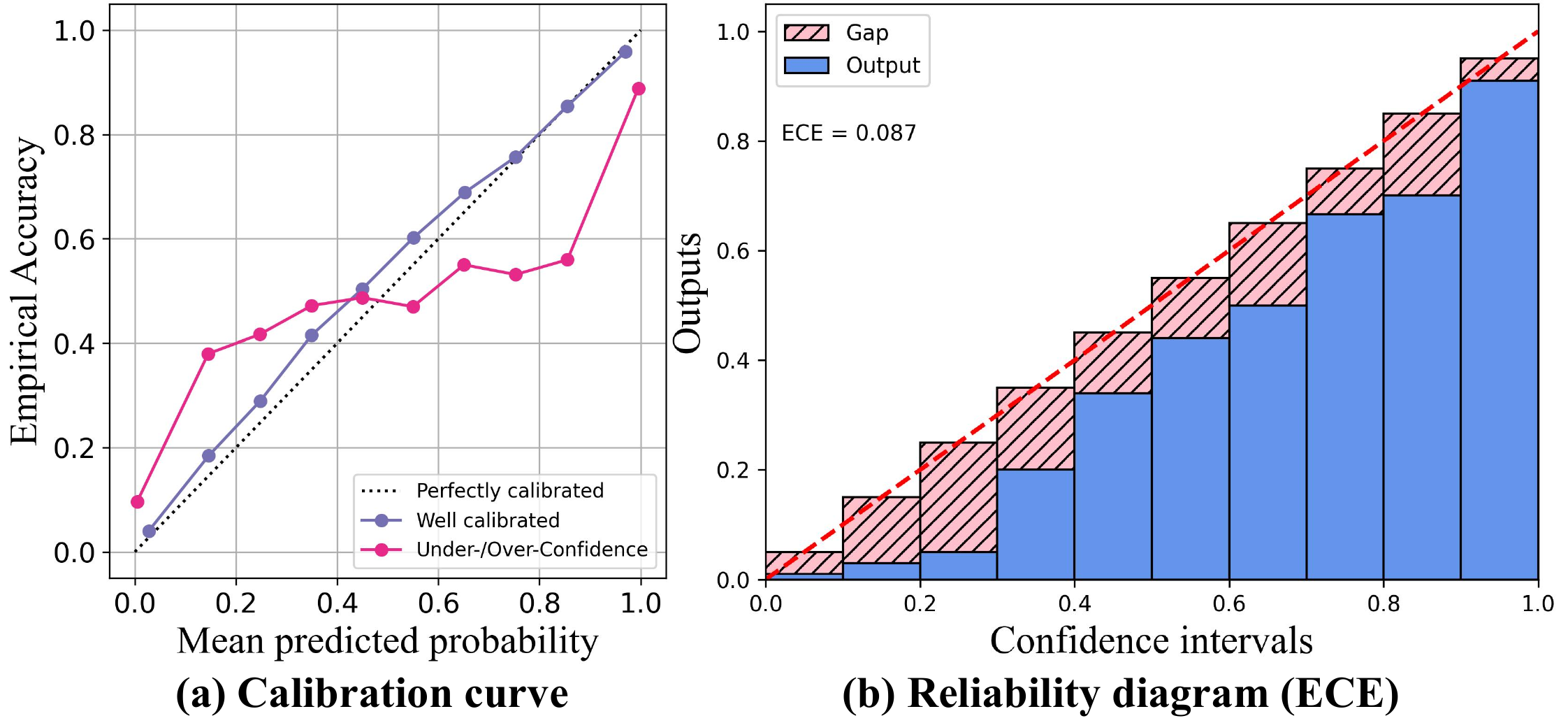}
    \caption{Illustration of uncertainty representation method based on calibration confidence. a) Calibration curve, the closer to the perfect curve, the better the confidence calibration. b) Reliability diagram combined with ECE calibration indicators.}
    \label{fig:Reliability_diagram}
\end{figure}

Calibration methods aim to correct the reliability of the uncertainty estimates provided by a model. The basic idea is to measure the accuracy of the predicted probabilities against the true probabilities. As shown in Fig. \ref{fig:Reliability_diagram}, if the predicted probabilities are well calibrated, then they will accurately reflect probabilities, and the model will be considered reliable. It is worth noting that the probability distribution can be obtained by a single forward pass of the network, also including some post-processing methods. We introduce below two commonly used methods of expressing uncertainty, as well as calibration methods in NLP research.

\textbf{Softmax Response.} Confidence method is a widely used uncertainty estimation method. It is based on the idea that the predicted category with the highest probability is the most likely category, and the uncertainty can be represented by Softmax Response (SR). There are also methods based on the difference (SD) between the top two values of the Softmax output, with a slight difference indicating uncertainty and a significant difference indicating confidence. Given an input ${\mathbf{x}^{(i)}}$, the uncertainty describe as:
\begin{equation}
    u_{SR}(\mathbf{x}^{(i)}) = 1-{\mathop{\mathrm{max}}_{c\in C}\;p(y^{(i)}=c|\mathbf{x}^{(i)})}.
\end{equation}
\begin{equation}
    u_{SD}(\mathbf{x}^{(i)}) = u_{SR}(\mathbf{x}^{(i)}) - \mathop{\mathop{\mathrm{max}}_{c_{2}\in C}}_{c_{2}\neq	c}\,p(y^{(i)}=c_{2}|\mathbf{x}^{(i)}).
\end{equation}

\textbf{Entropy-based.} Prediction entropy, which can be understood as the total uncertainty of the output distribution. When the certainty of the output distribution is higher, the entropy value is lower. Maximum entropy is reached when all outcomes have equal probability. In other words, entropy is highest value when the model is unable to distinguish between different outcomes. The predictive entropy at point $\mathbf{x}^{(i)}$ is equal to the conditional entropy of the output random variable $Y$:
\begin{equation}
    PE(\mathbf{x}^{(i)})=-\int p(y|\mathbf{x}^{(i)})\,{\mathrm{ln}}\,p(y|\mathbf{x}^{(i)})dy.
\end{equation}
In NLP research, the entropy value can represent a variety of uncertain information, such as the \textit{N-grams} word entropy value as a heuristic feature to select domain adaptation data \cite{ruder-plank-2017-learning}, the entropy value at the sentence level can represent semantic uncertainty \cite{kuhn2023semanticeq}, The composition of information entropy in NLP is diverse and complex, and these issues need to be considered when using entropy values.

Recently, the problem of overconfidence can arise in DNNs \cite{guo2017calibration}, often requiring calibration to obtain reliable confidence. A simple but effective calibration method is \textbf{Temperature scaling (TS)}, which was originally introduced into the machine learning community as a tool for knowledge distillation \cite{hinton2015distilling}. Specifically, TS helps introduces a temperature parameter $T > 0$ and generates a calibrated prediction vector by mapping:
\begin{equation}
    \mathbf{q}^{(i)} = \sigma_{SM}(\frac{\mathbf{z}^{(i)}}{T}),
\end{equation}
where $\sigma_{SM}(\mathbf{z})=e^{\mathbf{z}}/\sum_{k=1}^{K}e^{z_{k}}$ refers to the logits $\mathbf{z}$ pass the Softmax function. With $T=1,\mathbf{q}^{(i)}=\hat{\mathbf{p}}^{(i)}$, when $T>1$, the entropy of $\mathbf{q}^{(i)}$ increases, which helps reduce confidence $s^{(i)}$ and combat overconfidence, and larger $T$ will make the distribution flatter. Similarly, $T<1$ reduces entropy and increases confidence, making the distribution sharper to help with underconfidence predictions. The temperature parameter $T$ is trained with NLL on the validation set. TS acts as a post-processing calibration method, treating each prediction independently without explicitly taking contextual information into account. Therefore, it's important to consider whether token-level or sequence-level calibration is more appropriate for the specific NLP task at hand.

In addition, there are some implicit calibration methods to combat overconfidence, such as \textbf{Label smoothing} and \textbf{Re-weighting techniques} \cite{mukhoti2020calibrating,zhu2022boundary,xiao-etal-2022-uncertainty}. Label smoothing is a regularization technique that replaces the one-hot target vector $\mathbf{y}^{(i)}$ with a weighted combination of targets that have a uniform distribution. The amount of smoothing is controlled by an $\alpha$ parameter. In the $K\!-\!class$ setting, the true class represented by the value 1 is replaced by $1-\alpha$, and the other classes are assigned the value of $\frac{\alpha}{K-1}$. Label smoothing has been shown to have several benefits for the model predictions. It implicitly calibrates model predictions similar to TS and does not require post-operation quantification. Similarly, reweighting techniques such as Focal loss can also reasonably reduce the sharpness, and thus improve the calibration effect and performance \cite{lin2017focal, mukhoti2020calibrating}. By increasing the entropy of the target distribution, label smoothing and Focal loss prevent the model from producing extremely low-entropy predictions. 

\textbf{Conformal Prediction (CP)} \cite{vovk2005algorithmic} is a framework for constructing calibrated predictive models that provide a measure of confidence for each prediction. When CP is formulated in terms of a set of predictions $\mathcal{C}(X_{n+1})$, it provides finite-sample, distribution-free guarantees for events in which $\mathcal{C}$ contains $Y_{n+1}$. CP is based on hypothesis testing, where for a given input $x$ and possible output $y$, a statistical test is performed to accept or reject the null hypothesis that the pairing $(x, y)$ is correct. It can be seen as a post-hoc calibration method, which means that it does not require access to the training data during the calibration process. Instead, it uses the model's predictions on a validation set or a hold-out set to construct a set of prediction regions or intervals.

\textbf{Challenging:} \textit{How to solve the trade-off between accuracy and calibration?} As increasing the accuracy may result in overfitting and poor calibration, while increasing the calibration may result in underfitting and poor accuracy. Finding the optimal balance requires careful consideration of the model complexity and the application requirements. Furthermore, NLP tasks often involve making predictions at different levels of granularity, such as sentence-level, document-level, or token-level predictions. Uncertainty quantification of texts with different granularities is also worth thinking about.

\subsubsection{Sampling-based Methods} Sampling methods provide a representation of the target random variable, which can be divided into parametric and predictive sampling, from which implicit conditional distributions of the posterior model parameters can be derived. Bayesian variational inference approximates the posterior distribution of model parameters. It optimizes a variational objective function. This method estimates uncertainty in model predictions by sampling from the approximate posterior distribution. For instance, variational inference methods approximate intractable posterior distributions \cite{neal1992bayesian,blundell2015weight} by optimizing a family of tractable distributions. A commonly used method is MC dropout \cite{gal2016dropout}, which involves randomly dropping out some neurons in the network during testing. This technique results in a different prediction for each dropout configuration.
These methods are based on Markov Chain Monte Carlo and further extensions \cite{hinton1993keeping}. Furthermore, sampling-based methods also include ensemble methods. These methods involve training multiple models with different initializations or architectures and combining their predictions to make a final prediction.

\textbf{Monte Carlo dropout (MCD)} is the typical sampling method. 
Specifically, the MCD-based method performs $M$ stochastic forward passes with activated dropout, explicit ensembling to approximate integral Eq. \ref{posterior_bnn}:
\begin{equation}
    {{P}}({y}^{(i)}|\mathbf{x}^{(i)},\mathcal{D}) \approx \frac{1}{M}\sum_{i=1}^{M} {{P}}({y}^{(i)}|\mathbf{x}^{(i)},\hat{\boldsymbol{\theta}}^{(i)}
    ), \quad {\hat{\theta}}^{(i)}\sim {{q}}(\boldsymbol{\theta})
    \label{posterior_sample}
\end{equation}
Each ${{P}}({y}^{(i)}|\mathbf{x}^{(i)},\hat{\boldsymbol{\theta}}^{(i)})$ sampled from the distribution ${{q}}$ is a categorical distribution of class labels on a given input x. It visualizes as a point on the simplex \cite{malinin2018prior}.

Therefore, these methods construct an implicit conditional distribution simplex by sampling, and the characteristic is that whether a certain answer can be determined through the sharpness of the simplex, that is, the expression containing uncertainty. Given such a collection of distributions, it is expected that the entropy of the distribution $P$ will indicate the uncertainty of the prediction. It should be noted that although the uncertainty expressed by entropy is effective, it cannot be directly distinguished as aleatoric uncertainty or epistemic uncertainty. To address this, measures such as mutual information can be used to estimate prediction uncertainty due to epistemic uncertainty \cite{poole2019variational,yu2023learning}.

In addition, Bayesian active learning by disagreement (BALD; \cite{houlsby2011bayesian}) can be seen as a way to quantify uncertainty based on MCD:
\begin{equation}
    u_{\mathrm{BALD}} = -\sum_{k=1}^{K}\overline{p}_{c}\;{\mathrm{log}}\;\overline{p}_{c} + \frac{1}{M}\sum_{c, m}p_{c}^{t}\;{\mathrm{log}}\;p_{c}^{t}
\end{equation}

From a Bayesian perspective, MCD is a way of approximating Bayesian inference in neural networks \cite{gal2016dropout}. It can be understood as placing a prior distribution on the weights of the neural network, and then using dropout to sample from the posterior distribution. 

\textbf{Ensemble Methods}.
An alternative approach to Bayesian approximation involves creating ensembles of multiple independent deterministic neural networks. The technique involves training $M$ neural network classifiers, with different models randomly initialized and optimized individually, and combining their networks outputs to form a single classification function $f(x):X \rightarrow Y$, For example, this can be achieved by simply averaging the predictions of the members: 

\begin{equation}
    f(x) := \frac{1}{M} \sum_{i=j}^{M} f_{j}(x).
\end{equation}

The premise behind this is to leverage diverse perspectives to improve the generalization of the model, on the basis that a group of decision-makers is typically more effective than a solitary one. Ensembles have been found to be particularly effective in uncertainty estimation networks in neural systems. Moreover,to measure different sources of uncertainty, mutual information (MI; \cite{houlsby2011bayesian, poole2019variational}) between ensemble predictions and their parameters is often used in NLP models \cite{li2021multitaskdense,rainagales2022answer,andersenmaalej2022efficient}. Furthermore, MCD can be thought of as an ensemble technique that simulates different models using dropout \cite{lakshminarayanan2017simple}, wherein a set of independently trained networks with their own weights are generated. By aggregating the predictions from these networks, we can obtain more accurate and robust predictions. In addition, recent work has achieved \emph{``better''} uncertainty by combining the expressive power of ensemble methods with MCD \cite{pop2018deep}. It has been demonstrated that sampling-based methods exhibit inherent calibration and robustness to unknown data representations \cite{ovadia2019can}. 

\textbf{Challenging:} \textit{How to design methods to reduce time and computational cost}? A set of models with varying initializations has the potential to explore multiple modes of the parameter space \cite{fort2019deep}. Reducing sampling time and computational cost is a challenging task in sampling-based uncertainty estimation models. It is important to maintain the diversity of individual models while achieving these goals.

\subsubsection{Distribution-based Methods.}  
As shown in Fig. \ref{fig:uncer_model}, a distribution-based model is also a deterministic model, but requires a specific distribution for uncertainty modeling. In general, distribution-based models need to pass some prior/posterior assumptions to allow the model to learn distribution information during training. The following are two commonly used examples of distribution-based uncertainty modeling.

\textbf{Dirichlet-based Uncertainty Models (DBU)}
utilize logits and construct a Dirichlet distribution. Unlike standard (Softmax) neural networks, this model predicts the parameters of the Dirichlet distribution instead of point estimates. The density of the Dirichlet distribution is defined as
\begin{equation}
    \begin{aligned}
        {\rm{Dir}}(\mathbf{p}^{(i)}|\boldsymbol{\alpha}^{(i)})=\frac{1}{B(\boldsymbol{\alpha}^{(i)})}\prod_{c=1}^{C}p_{c}^{(\alpha^{(i)}_{c}-1)},
    \end{aligned}\label{dir_density}
\end{equation}
where $B(\boldsymbol{\alpha}^{(i)})$ is the $C$-dimensional multinomial beta function.

Specifically, the $P(y^{(i)}|\mathbf{x}^{(i)}, \boldsymbol{\theta})$ obtained by the neural network is a categorical distribution ${\mathrm{Cat}}(\mathbf{p}^{(i)})$, which is a point on the simplex, and the Dirichlet distribution is a
prior distribution over categorical distribution, which is parameterized by its concentration parameters, and the epistemic uncertainty on $\mathbf{x}^{(i)}$ can be expressed by ${q}^{(i)}={\mathrm{Dir}(\boldsymbol{\alpha}^{(i)})}$, so it can be interpreted as the distribution of the categorical distribution.
\begin{equation}
    {{p}^{(i)}_{c}}\!=\!\frac{{\alpha}^{(i)}_{c}}{\alpha^{(i)}_{0}}, \; \quad y^{(i)}\!=\!\mathop{\mathrm{arg\,max}}\limits_{c\in C}\left[{{p}^{(i)}_{c}}\right].
    \label{eq:p_predict}
\end{equation}

There are currently some studies using the Dirichlet distribution to quantitative uncertainty \cite{malinin2018prior, sensoy2018evidential, charpentier2020posterior,zhang2023ner}. The advantage is that the network can obtain uncertainty estimates once forward. In addition, this parameterization allows the calculation of closed-form classical uncertainty measures \cite{kopetzki2021evaluating}, such as the differential entropy of Dirichlet distribution $m^{(i)}_{\mathrm{diffE}}=H({\mathrm{Dir}(\boldsymbol{\alpha}^{(i)})})$ or mutual information $m^{(i)}_{\mathrm{MI}}=I({y^{(i)},(\boldsymbol{p}^{(i)})})$ \cite{malinin2018prior}.

Moreover, constructing distributions by modeling networks is also a deterministic approach. For example, modeling a Dirichlet distribution on class probabilities rather than the point estimate of a Softmax output \cite{malinin2018prior,sensoy2018evidential,charpentier2020posterior}.

\textbf{Gaussian-based Uncertainty Models (GBU)}. Parameterizing uncertainty information using the mean and variance of a Gaussian distribution is a common approach used in various fields. Specifically, referring to $\mu(\mathbf{x})$ and $\sigma(\mathbf{x})$ as functions parameterized by $W$, the output mean and standard deviation are computed for input $\mathbf{x}$, assuming $y \sim \mathcal{N}(\mu(\mathbf{x}),\sigma(\mathbf{x})^{2})$ for the data generating process in the regression setting, and for the logits vector $\mathbf{z}$ in the classification setting samples are then converted to probabilities using a Softmax operation. This process can be described as The process is described as:
\begin{equation}
\begin{aligned}
    &\mathbf{u} \sim \mathcal{N}(\boldsymbol{\mu}(\mathbf{x}),\boldsymbol{\sigma}(\mathbf{x})^{2}),\\
    &\boldsymbol{p} = {\mathrm{Softmax}}(\mathbf{u}),\\
    &y \sim {\mathrm{Categorical}}(\boldsymbol{p}).
\end{aligned}
\end{equation}

This method models uncertainty as a Gaussian distribution, where the mean represents the most likely value or prediction, and the variance represents the level of uncertainty or variability. By embedding words as Gaussian distributional potential functions in an infinite dimensional function space, Vilnis et al. \cite{vilnis2014word} not only maps word types to vectors, but also to soft regions in space, allowing uncertainty modeling of meaning and metaphor and providing a rich geometry of the latent space.

In NLP systems, Gaussian distributions and Mahalanobis distance
can be used together in uncertainty estimation techniques \cite{zhouetal2021contrastive,vazhentsevetal2022uncertainty}. Gaussian distributions are used to model variables, while Mahalanobis distance are used to compare simulated distributions with actual observed data and quantify the distance between them. For example, measure the Mahalanobis distance between a test instance and the closest class conditional Gaussian distribution to estimate uncertainty:

\begin{equation}
    u_{MD} = {\min\limits_{c\in C}}(h^{(i)} - \mu_{c})^{T}\Sigma^{-1}(h^{(i)} - \mu_{c}),
\end{equation}
where $h^{(i)}$ denotes the hidden representation of the $i-th$ sample.

Furthermore, the Bayesian neural network can approximate the distribution of the function by learning the parameter distribution of some models, and the Gaussian processes (GPs) non-parametric Bayesian model directly uses the function $f$ to approximate the distribution of the function \cite{williams2006gaussian}, which can measure the uncertainty of the model. Formally, a Gaussian process can be defined as a set of random variables in which any limited number of variables follow a joint Gaussian distribution. To fully specify GP, two functions are needed, the mean function $m(\mathbf{x})$ and the covariance function $k(\mathbf{x},\mathbf{x}')$. If a function $f$ is based on GPs distribution of these two functions, it can be defined as:

\begin{equation}
    \begin{aligned}
        &m(\mathbf{x}) = \mathbb{E}[f(\mathbf{x})] \\
        &k(\mathbf{x},\mathbf{x}') = \mathbb{E}[(f(\mathbf{x})-m(\mathbf{x}))(f(\mathbf{x}')-m(\mathbf{x}'))] \\
        &f(\mathbf{x})\sim \mathcal{GP}(m(\mathbf{x}), k(\mathbf{x},\mathbf{x}')), 
    \end{aligned}
\end{equation}
where $m(\mathbf{x})$ is the mean function, which is usually the 0 constant, and $k(\mathbf{x},\mathbf{x}')$ is the kernel or covariance function, which describes the covariance between values of $f$ at the different locations of $k(\mathbf{x}$ and $\mathbf{x}')$.

GPs are an alternative kernel-based framework that provides competitive results for point estimation \cite{shah2013investigation,beck2014joint,beck2016exploring}, and they explicitly epistemic uncertainty in data and predictions. This makes GPs ideal when well-calibrated uncertainty estimates are required. A commonly used approach to uncertainty estimation using GPs is the posterior predictive distribution. The posterior predictive distribution is the distribution of the predicted value of the output variable at any given input point, given the observed data. The mean of the posterior predictive distribution gives the predicted value, while the variance provides a measure of uncertainty. The larger the variance, the more uncertain the forecast. However, the time complexity of GPs increases with the size of the data, which makes it intractable in many practical applications.

\textbf{Challenging:} \textit{How to make reasonable assumptions about different distributions in distribution-based uncertainty estimation modeling?} Distribution-based uncertainty estimation modeling often requires making assumptions about the underlying distribution of the data, which can be challenging when dealing with complex or multi-modal data. Developing techniques for selecting appropriate distributions and assessing their validity is necessary to ensure the accuracy of the model and the reliability of the uncertainty estimates.

\subsection{Uncertainty Estimation Metrics}\label{ue_metrics}
\subsubsection{Calibration Metrics}
The calibration metric assesses the consistency of the classifier in producing reliable predictions. Fig. \ref{fig:Reliability_diagram} can be intuitively visualized by (a) Calibration curves and (b) Reliability diagram. This part focuses on the following metrics for evaluating the calibration of predictions produced by deep learning classifiers using uncertainty estimation techniques.


\textit{Expected Calibration Error (ECE)} \cite{guo2017calibration}. It denotes the expected calibration error, which aims to evaluate the expected difference between model prediction confidence and accuracy. The concrete formulation is as follows:
\begin{equation}
    \begin{aligned}
        ECE = \sum_{i=1}^{|B|} \frac{N_i}{N}\lvert{\mathrm{acc}(b_i)-\mathrm{conf}(b_i)}\rvert,
    \end{aligned}
    \label{eq:ece}
\end{equation}
where $b_i$ represents the $i$-th bin and $|B|$ represents the total number of bins. $N$ denotes the number of total samples. $N_{i}$ represents the number of samples in the $i$-th bin. $\mathrm{acc}(b_i)$ denotes the accuracy and $\mathrm{conf}(b_i)$ denotes the average of confidences in the $i$-th bin.

\textbf{Maximal Calibration Error (MCE)} \cite{guo2017calibration}. MCE represents the maximum deviation between model accuracy and prediction confidence. In programs that require a reliable measure of confidence, it is desirable to minimize the worst-case deviation between confidence and accuracy. MCE is similar to ECE in that this approximation involves bins:
\begin{equation}
    MCE = \mathop{\mathrm{max}}_{m \in \{1,...,M\}}|{\mathrm{acc}(b_m)-\mathrm{conf}(b_m)}|.
\end{equation}

\textbf{Brier Score (BS)}. The BS is a metric that evaluates the proximity of a model's predicted probability to the actual class probability, which is always 1. A desirable BS is near zero, as this indicates minimal deviation from the true class probabilities due to the squared differences \cite{assel2017brier}. BS is often written as the mean square prediction error (MSE) in regression tasks, defined as:
\begin{equation}
    BS = \frac{1}{N}\sum_{i=1}^{n}(y^{(i)}-\hat{y}^{(i)})^{2}.
\end{equation}

Both BS and mean squared error (MSE) involve calculating the squared difference between predicted and actual values. BS is often used to evaluate probabilistic predictions, while MSE is more commonly used to evaluate regression models.

\textbf{Negative Log-likelihood (NLL)}. The NLL is a widely accepted metric for evaluating the effectiveness of a probabilistic model \cite{quinonero2006evaluating}, and in the realm of deep learning it is commonly referred to as cross entropy loss. When presented with a probabilistic model and a set of $n$ samples, this measure can be used to gauge the model's calibration:
\begin{equation}
    NLL = -\sum_{i=1}^{n}{\mathrm{log}(\hat{\pi}(y^{(i)}|\mathbf{x}^{(i)}))}.
\end{equation}

Both NLL and Negative Log Predictive Density (NLPD) are calculated based on the negative logarithm of probability, and NLPD calculates the negative logarithm of the predicted density of new data given the training data and model.

Furthermore, some NLP research proposes to calibrate evaluation metrics to make them more suitable for evaluating specific tasks. For instance, Lin et al. \cite{lin2023on} proposes a new metric called Compositional Expected Calibration Error (CECE) in his Seq2Seq graph parsing work to measure the behavior of the model in predicting the structure of the combined graph, which evaluates the performance of the model on graph elements. This metric calibrates the case to better explain the behavior of the model in the combined graph structure and the distributional behavior of predicted graph structures under offset.

\subsubsection{Uncertainty Indication}
Apart from assessing confidence calibration, there are several other evaluation indicators for uncertainty. These metrics are primarily focused on constructing classification metrics by utilizing either uncertainty or confidence to determine the model's capability to differentiate between misclassified and outlier samples.

\textbf{The Area Under the Curve (AUC)}. AUC is a commonly used measure the area under the receiver-operator curve, the formulation is as follows:
\begin{equation}
    {AUC} = \frac{\sum_{t_0\in\mathcal{D}^0}\sum_{t_1\in\mathcal{D}^1}\mathbbm{1}[g(t_{0})<g(t_{1})]}{|\mathcal{D}^0|\cdot|\mathcal{D}^1|}\label{auc}
\end{equation}
where $\mathcal{D}^0$ is the set of negative examples, and $\mathcal{D}^1$ is the set of positive examples. $\mathbf{1}[g(t_{0})<g(t_{1})]$ denotes an indicator function which returns 1 if $g(t_{0})<g(t_{1})$ otherwise return 0.

Area Under the Receiver-Operator Characteristic Curve (AUROC), and the AUROC metric measures the probability that a randomly selected accurate response has a greater uncertainty score than a randomly selected inaccurate response. Higher values indicate superior performance, with an ideal uncertainty score of 1 and a value of 0.5 for a random uncertainty measure. Therefore, AUROC scores high when the model is confident about correct predictions but uncertain about incorrect ones. Depending on the axes and decision information used, AUC can produce varying score indicators, such as AUCPR (Area Under the Precision-Recall Curve) can be seen as an extension of the AUC score. However, AUC is based on the True Positive Rate (TPR) and False Positive Rate (FPR), while AUCPR is based on Precision and Recall.

\begin{figure}
    \centering
    \includegraphics[scale=0.38]{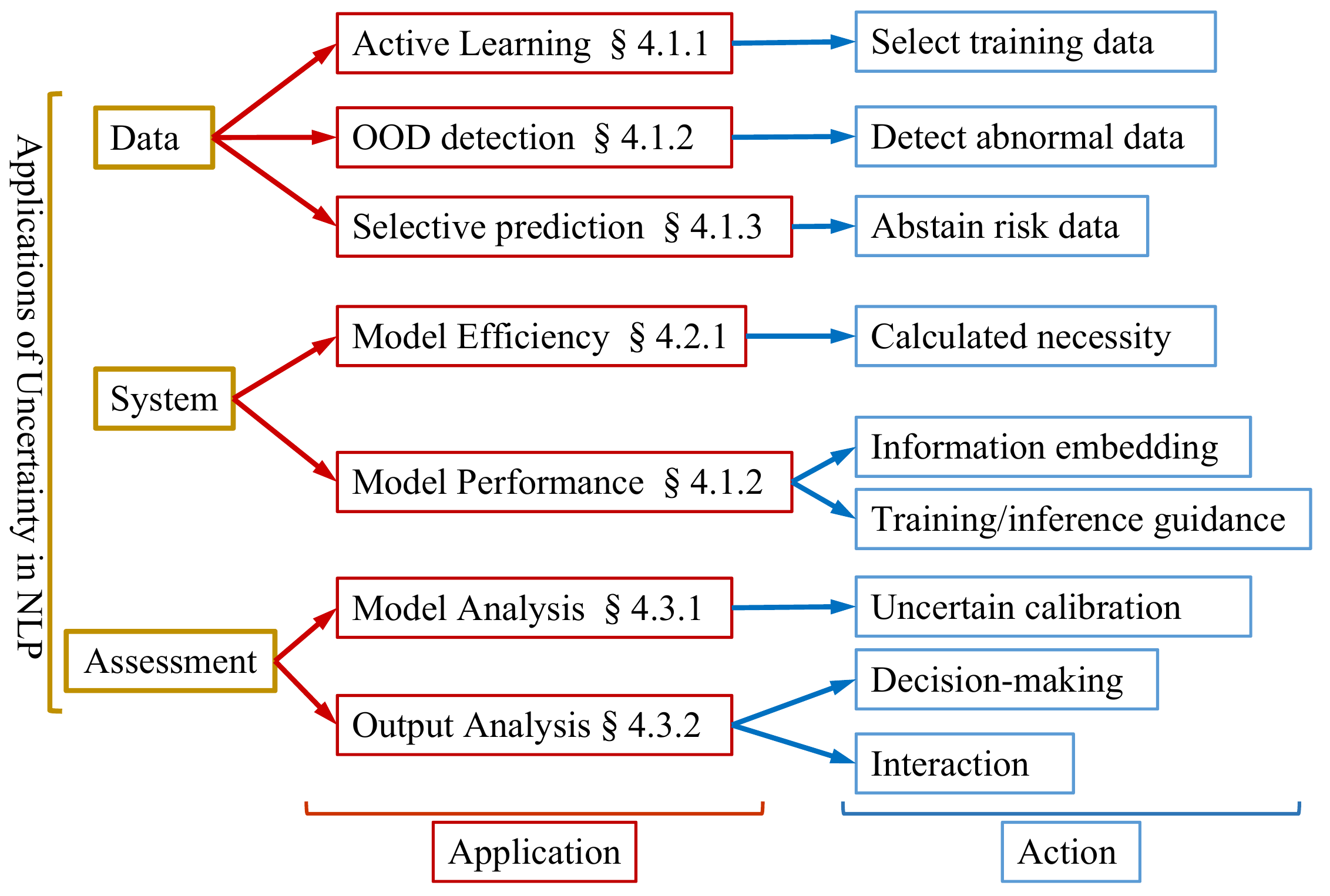}
    \caption{An overview of the application of uncertainty estimation to NLP systems. The red box indicates how the uncertainty is applied, and the blue box is the action taken using the uncertainty.}
    \label{fig:application}
\end{figure}

\textbf{The Area Under the Risk Coverage Curve (AUC-RCC)} \cite{Yaniv2010Selective}. AUC-RCC evaluates the quality of uncertainty estimation in terms of its ability to reject predictions with high uncertainty and avoid misclassification. AUC-RCC is based on the risk coverage curve, which plots the cumulative risk (or loss) as a function of the uncertainty level used for rejecting predictions.

\textbf{The Reversed Pair Proportion (RPP)} \cite{xinetal2021art}. Given a labeled dataset $D$ of size $n$, RPP is used to evaluate the distance of the uncertainty estimator $\tilde{u}$ from the ideal value. In order to minimize the AUC of RCC, the selective classifier should be intuitive for the correct classification output $\tilde{u}=0$ for the example of correct classification output, and $\tilde{u}=1$ for the wrong example. Therefore, the formula is defined as follows:
\begin{equation}
    RPP = \frac{1}{n^{2}}{\sum_{1 \leq i,j \leq n}^{n}\mathbbm{1}[\tilde{u}(x_{i})>\tilde{u}(x_{j}),l_{i}<l_{j}]},
\end{equation}
where $n^{2}$ in the denominator is used to normalize the values.

RPP measures the proportion of pairs of examples with an inverse confidence-error relationship, and an ideal confidence estimator would have an RPP value of 0. In contrast to AUROC, which measures overall discrimination, the RPP incorporates consistency of ranking.

\begin{figure}
    \centering
    \includegraphics[scale=0.45]{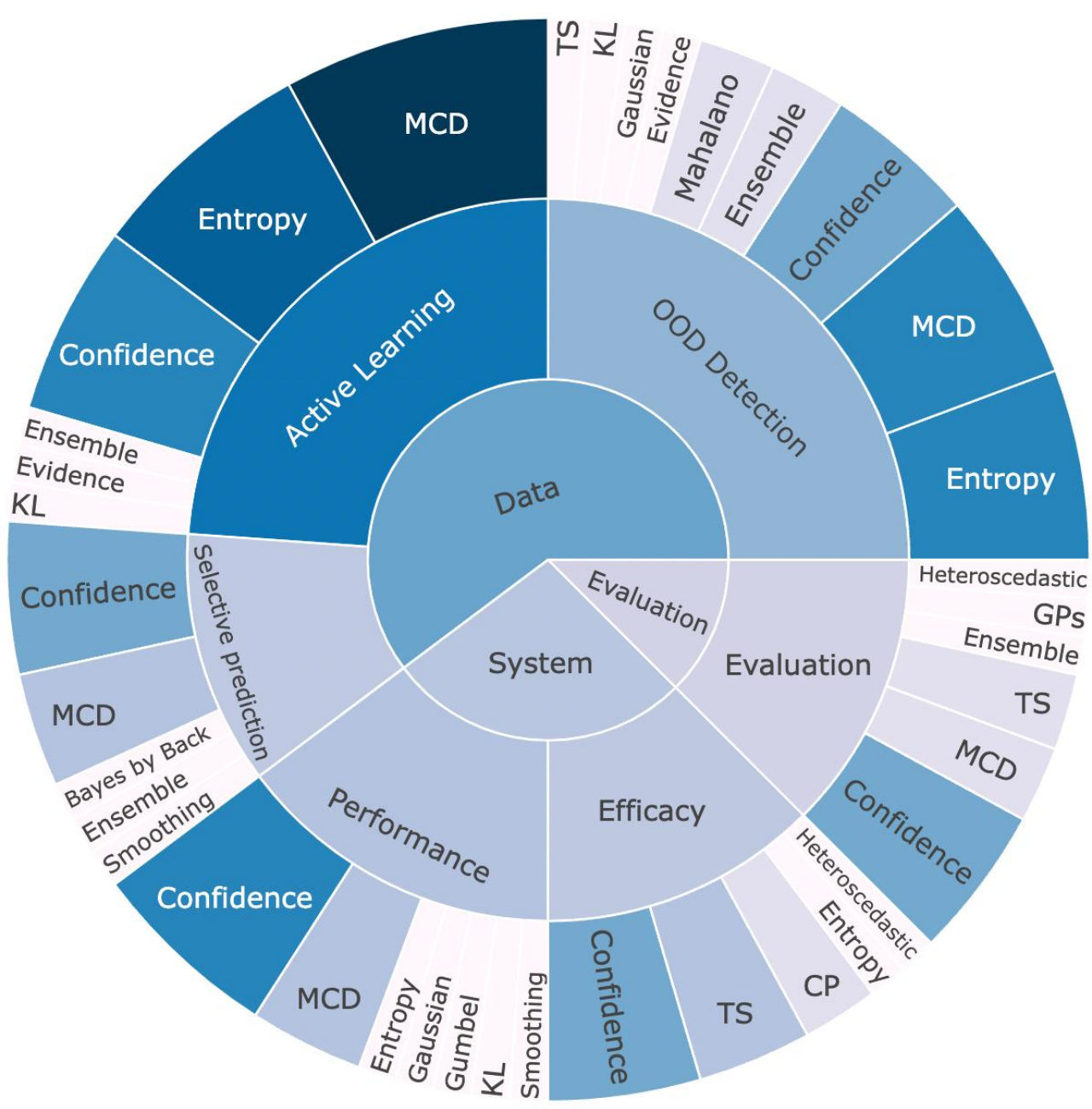}
    \caption{An overview of the applied classification of uncertainty. The inner circle represents the application, the middle circle represents the uncertainty modeling method used, and the outer circle refers to the specific uncertainty estimation method.}
    \label{fig:statistics_2}
\end{figure}

\section{Applications of Uncertainty in NLP}\label{APP_UE_NLP}
In this section, we review the applications of uncertainty estimation to NLP systems. As shown in Fig. \ref{fig:application}, the high-level perspective is mainly divided into three dimensions: \emph{data}, \emph{system}, and \emph{assessment}. The proportions of each type and its sub-types are further counted, depicted in Fig. \ref{fig:statistics_2}.

\subsection{Data Filter and Action Guidance}
\label{sec:4.1}
In the \emph{data} dimension, we summarize the application of uncertainty as filtering and guidance. It mainly involves three primary aspects, discussed separately below.

\begin{table*}
    \renewcommand{\arraystretch}{1.3}
    \caption{Application of Uncertainty-Based Data Correlation in NLP, and some task/method names are represented as abbreviations.}
    \label{tab:app_data}
    \centering
    \begin{tabular}{lllll}
        \toprule
        Application&Paradigm&Tasks Types&UE\_Methods &Metrics\\
        \midrule
        Active Learning &Class. & \makecell[l]{TC\cite{zhu2008active,yuanetal2020cold,eindor2020active};\\\cite{yu2022actune,margatina2021active,ru2020active,NEURIPS2020_f23d125d};\\ 
        NLI\cite{lei2022uncertainty};\\
        SA\cite{margatina2021active}.} &\makecell[l]{Confidence\cite{yuanetal2020cold,eindor2020active,yu2022actune};\\
        Entropy\cite{zhu2008active,ru2020active};\\ KL\cite{margatina2021active};\\
        MCD\cite{eindor2020active,yu2022actune,NEURIPS2020_f23d125d}; \\
        Ensemble\cite{eindor2020active};\\ Evidence\cite{lei2022uncertainty}.}&\makecell[l]{Accuracy \cite{zhu2008active,yuanetal2020cold,eindor2020active};\\{\cite{yu2022actune,margatina2021active,NEURIPS2020_f23d125d,lei2022uncertainty}};\\ \\F1{\cite{yuanetal2020cold,eindor2020active,lei2022uncertainty}}.}\\
        
        Active Learning &SeqLab & \makecell[l]{NER\cite{siddhantlipton2018deep,shelmanov2021active,chaudhary2019little},\\\cite{liu2022ltp,ru2020active}.} &\makecell[l]{Confidence\cite{liu2022ltp};\\ Entropy\cite{chaudhary2019little,ru2020active};\\ MCD\cite{siddhantlipton2018deep,shelmanov2021active}.} &\makecell[l]{F1\cite{siddhantlipton2018deep,chaudhary2019little,shelmanov2021active,ru2020active};\\ Accuracy\cite{siddhantlipton2018deep,liu2022ltp}.}\\

        Active Learning &Generative & \makecell[l]{QA\cite{lyu2020you}; \\ AS\cite{gidiotis2022trust,xu2020understanding}.} &\makecell[l]{Confidence\cite{xu2020understanding};\\ MCD\cite{lyu2020you,gidiotis2022trust}.} &\makecell[l]{ROUGE\cite{xu2020understanding};\\$\rm{BLEU_{Var}}$\cite{lyu2020you,xu2020understanding,gidiotis2022trust}.}\\

        OOD detection &Class &\makecell[l]{SA\cite{hendrycks2020pretrained,yu2023learning,zhouetal2021contrastive};\\ SLU{\cite{shen2021enhancing,zhouetal2021contrastive}};\\
        NLI\cite{desai2020calibration};\\ TC\cite{zhang2019mitigating,desai2020calibration,he2020towards},\\\cite{margatina2021active,Hu2021URTX,vazhentsevetal2022uncertainty}.} &\makecell[l]{Confidence\cite{zhang2019mitigating,desai2020calibration,hendrycks2020pretrained,zhouetal2021contrastive};\\ Entropy\cite{yu2023learning,shen2021enhancing};\\ TS\cite{shen2021enhancing};\\ KL\cite{margatina2021active};\\ Mahalanobis\cite{zhouetal2021contrastive,vazhentsevetal2022uncertainty};\\ Evidence\cite{Hu2021URTX};\\
        MCD\cite{he2020towards};\\ Ensemble\cite{he2020towards};\\
        Gaussian\cite{shen2021enhancing}.}&\makecell[l]{ Accuracy\cite{desai2020calibration,he2020towards,hendrycks2020pretrained};\\\cite{margatina2021active,yu2023learning,shen2021enhancing};\\ F1,\cite{zhang2019mitigating,he2020towards,vazhentsevetal2022uncertainty};\\
        ECE\cite{desai2020calibration};\\ AUROC\cite{Hu2021URTX,zhouetal2021contrastive,shen2021enhancing};\\ AUPR\cite{Hu2021URTX,shen2021enhancing,vazhentsevetal2022uncertainty};\\
        FAR95\cite{hendrycks2020pretrained,zhouetal2021contrastive,shen2021enhancing};\\
        AUCRCC\cite{vazhentsevetal2022uncertainty},\\
        FPR90\cite{Hu2021URTX}.}\\
        
        OOD detection &SeqLab &NER\cite{vazhentsevetal2022uncertainty,yu2023learning,zhang2023ner}. &\makecell[l]{Entropy\cite{yu2023learning};\\
    Evidence\cite{zhang2023ner}.}&\makecell[l]{F1\cite{vazhentsevetal2022uncertainty,yu2023learning,zhang2023ner};\\AUROC \cite{zhang2023ner}.}\\
        
        OOD detection &Generative &\makecell[l]{LM\cite{yu2023learning};\\ MT\cite{xiao2020wat,wu2021uncertainty};\\ QA\cite{li2021multitaskdense}.} &\makecell[l]{MCD\cite{xiao2020wat,wu2021uncertainty};\\ Entropy\cite{yu2023learning,shen2021enhancing};\\ Ensemble\cite{li2021multitaskdense}.}&\makecell[l]{Perplexity\cite{yu2023learning};\\$\rm{BLEU_{Var}}$\cite{xiao2020wat,wu2021uncertainty};\\
        Accuracy\cite{li2021multitaskdense}.}\\

        Selective prediction&Generative &\makecell[l]{QA\cite{kamath-etal-2020-selective,gargmoschitti2021will,varshneyetal2022investigating}.} &\makecell[l]{Confidence\cite{gargmoschitti2021will,varshneyetal2022investigating};\\ Smoothing\cite{varshneyetal2022investigating};\\ MCD\cite{varshneyetal2022investigating,kamath-etal-2020-selective};\\}&\makecell[l]{AUROC\cite{varshneyetal2022investigating,kamath-etal-2020-selective}.\\Accuracy\cite{kamath-etal-2020-selective};\\ Precision/Recall \cite{gargmoschitti2021will}.}\\ 

        Selective prediction&Class &\makecell[l]{Duplicate Detection\cite{gargmoschitti2021will,varshney2022towards};\\NLI\cite{gargmoschitti2021will,varshney2022towards};\\SA\cite{hendrycks2017a};\\TC\cite{andersenmaalej2022efficient}.} &\makecell[l]{Confidence\cite{hendrycks2017a,varshney2022towards};\\ MCD\cite{andersenmaalej2022efficient};\\ Ensemble\cite{andersenmaalej2022efficient};\\
        Bayes by Backprob\cite{andersenmaalej2022efficient}.}&\makecell[l]{Accuracy\cite{varshney2022towards,andersenmaalej2022efficient};\\
        F1\cite{andersenmaalej2022efficient};\\ AUROC\cite{hendrycks2017a,varshney2022towards,andersenmaalej2022efficient};\\
        AUPR\cite{hendrycks2017a}.}\\
        \bottomrule
        \end{tabular}
\end{table*}

\subsubsection{Active Learning}
Active learning (AL) is a machine learning technique that reduces the amount of labeled data required for training by selecting the most informative unlabeled data for annotation. This is based on the idea that not all data points are equally useful during learning. Uncertainty-based AL is a popular approach that assumes that data with higher uncertainty is more informative and likely to enhance model performance. As shown in Table \ref{tab:app_data}, uncertainty-based AL has been applied to various NLP tasks, and performance metrics (e.g., Accuracy, F1) are mainly used for AL evaluation. As shown in Fig. \ref{fig:statistics_2}, AL mainly uses the MCD based on sampling, and the confidence and entropy method based on calibration, focusing more on the quality of data filtered by the teacher model and the overall performance of the student system. 

Benefiting from the development of NLP's pre-training model, uncertainty-based AL can effectively reduce the labeling cost and improve model performance. For example, self-supervised language modeling and uncertainty filtering data can be used for AL in the cold-start setting, effectively reducing labeling costs and improving model performance \cite{yuanetal2020cold}. Ein-Dor et al. \cite{eindor2020active} combine BERT with the uncertainty estimation in the study of text classification AL to achieve good results in data screening of minority classes. In addition, pre-trained models allow us to improve data filtering capabilities in few-shot learning. Mukherjee et al. \cite{NEURIPS2020_f23d125d} consider fine-tuning DNNs with limited labeled data. This work develops two strategies for AL data filtering: filtering hard data with high uncertainty and filtering soft data with low uncertainty improves the performance of PLMs in few-shot learning scenarios. Combining pre-trained models with uncertainty-based active learning improves model performance by leveraging pre-existing knowledge and selecting informative examples for fine-tuning. 

Moreover, data can be filtered using uncertainty and additional information.
For example, ACTUNE \cite{yu2022actune} proposes uncertainty-based data filtering and guidance, allowing switching between data filtering and model training. This approach chooses highly deterministic unlabeled samples as actively labeled, while low-uncertainty samples are selected for model self-training. AL usually needs to traverse all unlabeled data to find informative unlabeled samples, which are always close to the decision boundary with large uncertainty. AUSDS \cite{ru2020active} annotates unlabeled text samples with high uncertainty through adversarial attacks, which significantly compresses the search space and protects the decision boundary from drastic changes.

Since AL can select samples with large uncertainty, it is natural to imagine applying uncertainty-based AL to multiple domains in the NLP. For example, Lyu et al. \cite{lyu2020you} identify language as a key factor in the difference between data inside and outside the distribution in QA. They further propose a data selection strategy for AL based on an uncertainty measure developed using deterministic sequence probability and MCD sequence probability. This approach was shown to be more effective in selecting uncertain data for multilingual QA. Information in real-world scenarios is often multi-domain and rich in unlabeled data. Siddhant et al. \cite{siddhantlipton2018deep} point out that classical uncertainty estimation can only account for arbitrary uncertainty (e.g., entropy or confidence). In contrast, using MCD+ BALD and Backprop+BALD as two uncertainty estimation methods can achieve excellent results in the three NLP tasks of emotion classification, NER, and semantic role labeling.

However, there are also some challenges and limitations of uncertainty-based AL, such as how to measure uncertainty effectively, how to trade off exploration and exploitation, and how to deal with the class imbalance and data diversity. This section provides a comprehensive review of existing work on uncertainty-based AL in NLP, covering data selection strategies, NLP issues involved, and uncertainty estimation methods.

\subsubsection{OOD/Outlier Detection}
Out-of-distribution (OOD)/outlier detection is an important application of uncertainty estimation in NLP. It helps to distinguish in-distribution examples from OOD/outlier examples. As shown in Fig. \ref{fig:statistics_2}, detecting OOD data in NLP covers three paradigms of uncertainty estimation. The goal is to identify examples that are different from the training data and the model has not been seen before. Metrics used to detect system performance include task performance metrics and uncertainty classification metrics. The task performance metric evaluates the model's ability to correctly handle the corresponding task, while the uncertainty classification metric measures the uncertainty quality of the model's predictions. We summarize the types of detection data into three ones: unknown domain data, misinformation data, and misclassification data.

\textbf{Unknown Domain Data Detection}. \; Domain shift in NLP can be of different types such as background shift (same task, but style/domain change) and semantic shift (unseen labels) \cite{hendrycks2019deep}. For example, traditional data selection based on testing domain knowledge often fails in unknown domains such as patents and tweets. Thus, unknown domain data detection is an important field of outlier data, including multi-domain intersection, open domain, etc. Uncertainty-based detection techniques can help solve these problems to improve the reliability and usability of NLP models in real scenarios. The corresponding NLP tasks have applications. For instance, MULTIUAT \cite{wu2021uncertainty} addresses the challenge of learning multilingual and multi-domain translation models due to heterogeneous and imbalanced data, which makes the model converge inconsistently over different corpora. 
Li et al. \cite{li2021multitaskdense} perform multi-task-intensive retrieval in open-domain QA, exploiting epistemic uncertainty to deal with corpus inconsistencies.
Yu et al. \cite{yu2023learning} show that maximizing the uncertainty of training data using entropy can enhance prediction accuracy on unseen domains and outperform CNN, BERT, and Transformer baselines, even without knowledge of unknown domains.

Different domains have various distributions. Therefore, some previous works apply distribution-based uncertainty estimation for OOD detection. Zhou et al. \cite{zhouetal2021contrastive} propose an unsupervised unknown domain detection method using a contrastive learning framework. They fine-tune Transformers with a contrastive loss to improve representation compactness. They further use Mahalanobis distance and Gaussian distribution in the penultimate layer to accurately detect OOD instances. Hu et al. \cite{Hu2021URTX} exploit uncertainty in OOD detection for text classification tasks using evidential neural networks based on Dirichlet distribution. They propose a framework that adopts auxiliary outliers and pseudo off-manifold samples to train the model with prior knowledge of a certain class, which has high vacuity for OOD samples. In addition, PLMs may affect OOD detection performance. Hendrycks et al. \cite{hendrycks2020pretrained} systematically study the OOD robustness of pre-trained Transformers on various models. To measure OOD detection performance, they use the negative prediction confidence as the outlier supervision score and show that the pre-trained Transformer performs better than previous models both in generalizing to OOD examples and in detecting OOD examples.

\textbf{Misinformation Data Detection}. \; The proliferation of misinformation on social media platforms is a major concern for society. One approach to developing uncertainty-based methods and preventing misinformation is to use uncertainty-based methods to detect and prevent misinformation and rumor spreading. To name a few,  
Zhang et al. \cite{Zhang2019Reply} conduct research on uncertainty estimation methods, specifically Bayesian deep learning and Evidence Lower Bound (ELBO) objective function for rumor detection.
Kochkina et al. \cite{kochkinaliakata2020estimating} research error/outlier detection for automatic rumor verification. They propose two uncertainty-based instance rejection methods using aleatoric and epistemic uncertainty estimates obtained through confidence-based and MCD-based methods. This research aims to solve the problem of resolving rumors circulating online by prioritizing difficult instances for human fact-checkers and interpreting model performance during rumor unfolding. Furthermore, Feng et al. \cite{fengetal2020none} study \emph{``none of the above''} (NOTA) detection tasks in dialogue systems, i.e. the case where no correct response (i.e. ground truth) exists in the candidate set. To this end, the paper utilizes MCD as an uncertainty estimation method to measure the ability to capture uncertainty for end-to-end retrieval models. Overall, the speed at which misinformation spreads on social media makes manual verification difficult, and these tools can help detect and mitigate the impact of misinformation, thereby reducing its potential negative impact on the public.

\textbf{Misclassification Data Detection}. \; To detect OOD, epistemic uncertainty is caused by limited training data or model structure. However, misclassification detection also requires modeling aleatoric uncertainty caused by noise and ambiguity in data \cite{mukhoti2021deterministic}.
Hendrycks et al. \cite{hendrycks2017a} study uncertainty estimation such as confidence-based and Softmax prediction probability for sentiment classification. The goal is to indicate when classifiers are likely to make mistakes to increase their adoption and prevent accidents. The authors provide a misclassification detection baseline for neural networks that outperforms MaxProb for selective prediction.
Vazhentsev et al. \cite{vazhentsevetal2022uncertainty} investigate the usage of uncertainty estimation for Transformer-based NER and text misclassification, and add an experimental setup given OOD data. The research proposes two computationally efficient modifications, including Mahalanobis Distance with a spectral-normalized network, which approach or outperform computationally intensive approaches.

However, \textit{how to determine the OOD boundary?} One of the main reasons for OOD boundary is the limited input information available to NLP models, as they typically rely on a limited set of features extracted from the input text. This can lead to many models that are well-calibrated or achieve excellent uncertainty estimation in IDs, but poor results in OOD samples \cite{kadavath2022largeModelKnow}. One option for finding the boundary between ID and OOD data is to train a supervised classifier on both ID and OOD data \cite{malinin2018prior,hendrycks2019deep}. Nevertheless, collecting a representative set of OOD data may not be practical due to the infinite compositionality of languages, and selecting an arbitrary subset may introduce selection bias and limit the generalizability of the model to unseen OOD data. Alternatively, Ryu et al. \cite{RYU201726,ryu2018domain} propose using generative models, such as autoencoders and GANs \cite{goodfellow2014generative}, to capture the ID data distribution and differentiate between ID and OOD data based on reconstruction error or likelihood. Other approaches, such as meta-learning for OOD detection and generating pseudo-OOD data \cite{tan2019domain}, also provide decision boundary between ID and OOD \cite{lee2018training}, but all require additional data or training procedures beyond the task, which may result in significant data collection work or inference overhead.

\begin{table*}
    \renewcommand{\arraystretch}{1.3}
    \caption{Uncertainty-based applications related to efficiency and performance improvement in NLP systems. }
    \label{tab:app_systems}
    \centering
    \begin{tabular}{lllll}
        \toprule
        Application&Paradigm&Tasks Types&UE\_Methods &Metrics\\
        \midrule
        Efficacy &Class&\makecell[l]{GLUE\cite{geng2021romebert,xinetal2021berxit};\\
        SA\cite{zhou2020bert,schwartzetal2020right,schusteretal2021consistent};\\
        NLI\cite{schwartzetal2020right};\\
        Fact verification\cite{schusteretal2021consistent};\\
        Topic classification\cite{schwartzetal2020right,schusteretal2021consistent};\\ Sentence classification\cite{liu-etal-2020-fastbert};\\Shopping review\cite{liu-etal-2020-fastbert};\\Sentences-matching\cite{liu-etal-2020-fastbert}.}
        &\makecell[l]{Confidence\cite{schusteretal2021consistent};\\ \\
        Entropy\cite{geng2021romebert};\\ \\
        TS\cite{schwartzetal2020right,schusteretal2021consistent};\\ \\
        CP\cite{schusteretal2021consistent}.} &\makecell[l]{ Accuaracy\cite{geng2021romebert,liu-etal-2020-fastbert};\\\cite{zhou2020bert,schwartzetal2020right,schusteretal2021consistent};\\
        Consistency\cite{schusteretal2021consistent};\\
        Layers\cite{schusteretal2021consistent};\\
        Time\%\cite{geng2021romebert};\\
        Speedup\cite{liu-etal-2020-fastbert,zhou2020bert,schusteretal2021consistent}.}\\
        
        Efficacy &Regression &\makecell[l]{GLUE\cite{zhou2020bert};\\ Similarity\cite{xinetal2021berxit,schusteretal2021consistent}.} &\makecell[l]{Confidence\cite{xinetal2021berxit,schusteretal2021consistent};\\ TS\cite{schusteretal2021consistent};\\
        CP\cite{schusteretal2021consistent}.} &\makecell[l]{Speedup\cite{zhou2020bert};\\
        Relative scores\cite{xinetal2021berxit}.}\\
        
        Efficacy &Generative&\makecell[l]{Text generation\cite{schuster2022confident}.}&\makecell[l]{Confidence\cite{schuster2022confident}.} &\makecell[l]{BLEU\cite{schuster2022confident};\\ ROUGE\cite{schuster2022confident};\\
        F1\cite{schuster2022confident};\\
        Layers\cite{schuster2022confident}.}\\

        Performance & Class &\makecell[l]{TC\cite{pei2022transformer};\\ \\Fake News Detection\cite{wei2022uncertainty}.} &\makecell[l]{Gumbel\cite{pei2022transformer};\\MCD\cite{pei2022transformer};\\Ensemble\cite{pei2022transformer};\\Gaussian\cite{wei2022uncertainty}.}&\makecell[l]{F1\cite{wei2022uncertainty};\\ Accuaracy\cite{pei2022transformer,wei2022uncertainty}.}\\
        
        Performance & Structure &KGs \cite{chen2019embedding,chen2021probabilistic,boutouhami2020uncertain}. &\makecell[l]{Gumbel\cite{chen2021probabilistic};\\
        Confidence\cite{chen2019embedding,boutouhami2020uncertain}.} &\makecell[l]{BS\cite{chen2019embedding, chen2021probabilistic,boutouhami2020uncertain};\\ MAE\cite{chen2019embedding,chen2021probabilistic};\\
        Accuaracy\cite{chen2019embedding};\\
        F1\cite{chen2019embedding}.}\\

        Performance &SeqLab &NER\cite{guietal2020uncertainty,zhu2022boundary,li2022rethinking}. &\makecell[l]{MCD\cite{guietal2020uncertainty};\\ 
        Entropy\cite{li2022rethinking};\\ Smoothing\cite{zhu2022boundary}.} &\makecell[l]{F1\cite{guietal2020uncertainty,zhu2022boundary,li2022rethinking};\\
        ECE\cite{zhu2022boundary}.}\\
        
        Performance &Generative&\makecell[l]{MT\cite{wang2019improving,zhou2020uncertainty,wei2020uncertainty};\\ \\QA\cite{zhang2021knowing}.}&\makecell[l]{
        Confidence\cite{zhang2021knowing,wei2020uncertainty};\\
        KL\cite{wei2020uncertainty};\\
        Norma\_Confidence\cite{zhou2020uncertainty};\\ MCD\cite{wang2019improving,zhou2020uncertainty}.}
        &\makecell[l]{BLEU\cite{wang2019improving,zhou2020uncertainty,wei2020uncertainty};\\
        Accuaracy\cite{zhang2021knowing};\\
        AUROC\cite{zhang2021knowing}.}\\

        \bottomrule
        \end{tabular}
\end{table*}

\subsubsection{Selective Prediction}
Selective prediction refers to the ability of AI systems to abstain from making predictions when faced with novel inputs that differ from their training data distribution. This is essential for improving the reliability of NLP systems in real-world safety-critical domains like biomedical and autonomous robots, where incorrect predictions can have serious consequences. Typically, $c$ contains a prediction confidence estimator $\tilde{c}$ and a threshold $\tau$ that controls the level of abstention:
\begin{equation}
    c(x) = \mathbbm{1}[\tilde{c}(x)>\tau]
\end{equation}

For dataset $\mathcal{D}$, the \textit{coverage} of the threshold $\tau$ corresponds to the fraction of answered instances (where $\tilde{c}(x)>\tau$), and the \textit{risk} is the error of these answered instances. 

Selective predictive systems make a trade-off between \textit{coverage} and \textit{risk}. As shown in Table \ref{tab:app_data}, QA systems for medical domains require high precision and discard questions that will not be answered by the QA system. This presents a cost-saving opportunity \cite{gargmoschitti2021will}. In addition, the example of a medical diagnosis model usually provides high-confidence classifications even when requires human intervention. The failure of classifiers indicates when they are likely mistaken, limiting their adoption or causing serious accidents. The application of selective prediction in NLP is crucial for enabling the reliable deployment of NLP systems in real-world applications.

Recently, selective prediction in NLP has received attention. Andersen et al. \cite{andersenmaalej2022efficient} study MCD for selectivity prediction in text classification. A semi-automatic text classification framework that minimizes unreliable and error-prone classifications by explicitly modeling uncertainty. 
Varshney et al. \cite{varshney2022towards} handle NLI tasks based on ID and OOD datasets, allowing systems to avoid making predictions when they might go wrong, improving their reliability in safety-critical domains.
Furthermore, Varshney et al. \cite{varshneyetal2022investigating} investigate selective prediction in NLP using 17 datasets covering NLI, duplicate detection, and QA tasks. In BERT-based experiments, the results show that these methods outperform the MaxProb method when comparing whether the MCD and label smoothing can improve the performance of selectivity prediction.
Garg et al. \cite{gargmoschitti2021will} conduct research on filtering out unanswered questions in medical QA systems. They focus on Transformer-based question models to improve answer models and estimate uncertainty. This research shows that confidence scores for answers can be approximated from question text alone without requiring answers. By filtering out error-prone problems, the improved QA model improves the reliability of real-world NLP applications.
Kamath et al. \cite{kamath-etal-2020-selective} explore selectivity prediction in QA tasks under domain shift. Here TS is used to calibrate confidence scores, which enables the model to discard answers when necessary to account for distributions different from the training data.

Selective prediction in NLP involves using uncertainty to determine which data to process, resulting in higher accuracy on the answered subset. However, determining how to rank examples according to confidence metrics and balancing the trade-offs between risk and coverage are still considerations in selective prediction.

\subsection{NLP System Efficiency and Performance}
The framework of an NLP system is usually multi-layer or multi-module. Uncertainty estimation can be exploited for each component to improve the computational efficiency or performance of the network. Although performances could be also improved by better dealing with data in Sec \ref{sec:4.1}, here in this section, we review from the perspective to leverage uncertainty in the interior computation of NLP systems and models. As presented in Fig. \ref{fig:application}, uncertainty is introduced for deciding the necessity of calculation. And it promotes performance by providing more information or guidance. 

\begin{figure}
    \centering
    \includegraphics[scale=0.4]{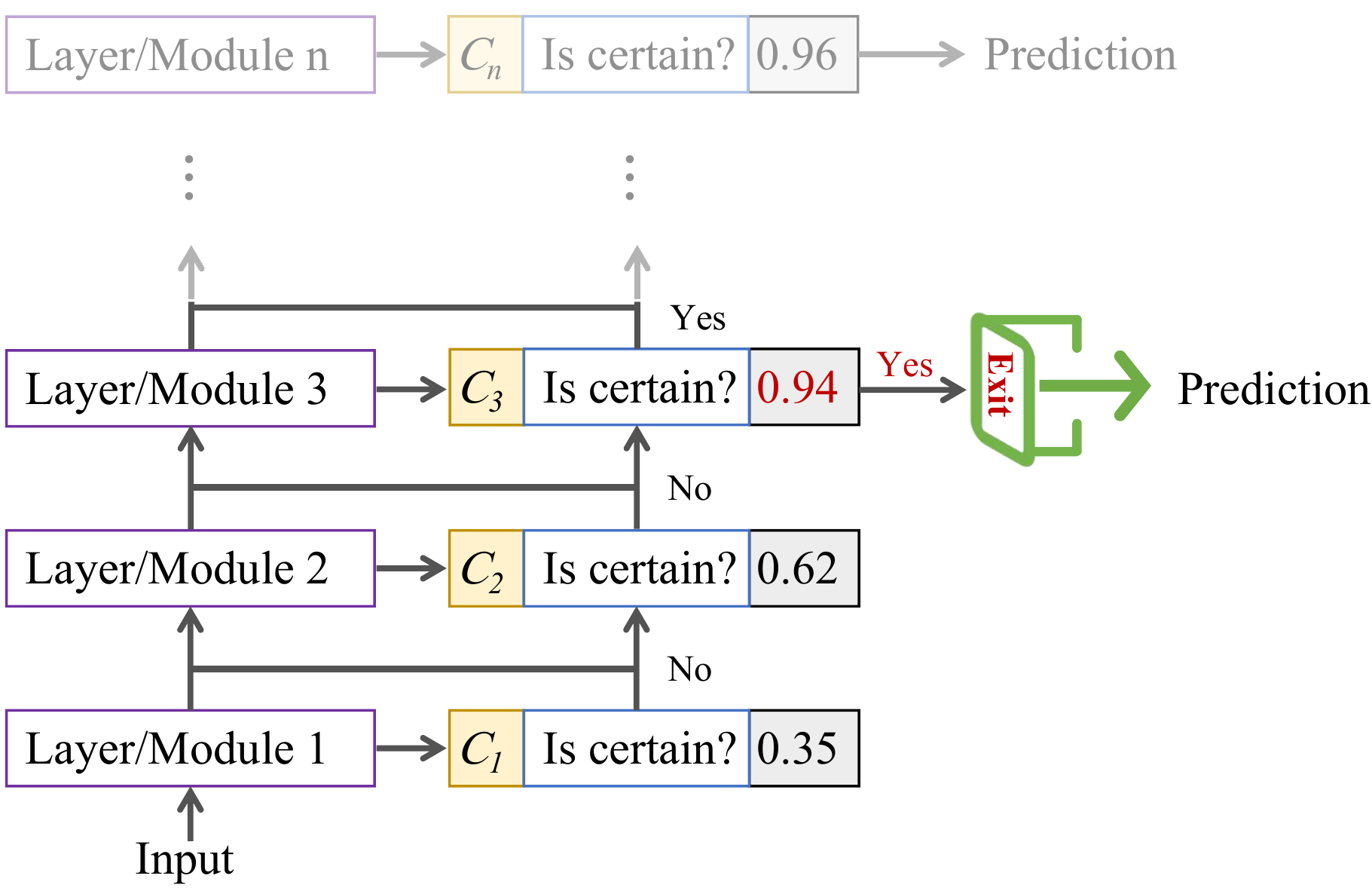}
    \caption{Illustration of the early exit method based on uncertainty. Common layers/modules such as Transformer, $C_{i}$ are usually a classifier, determining whether the early exit is determined by its confidence.}
    \label{fig:early_exit}
\end{figure}

\subsubsection{Uncertainty and Efficiency}
PLMs have impressive performance but require substantial computational resources, which limits their use in real-world applications \cite{schwartzetal2020right, schusteretal2021consistent, xinetal2021berxit}. In particular, autoregressive decoding using a full stack of Transformer layers for each output token is computationally expensive, and this approach is commonly used in NLP \cite{schuster2022confident}. This limitation makes it difficult where fast inference and low computational cost are essential. Additionally, over-parameterization of PLMs can lead to overthinking in decision-making, where the model may focus on complex or irrelevant features in later layers instead of relying on simpler features from earlier layers that generalize better \cite{zhou2020bert}.
As demonstrated in Fig. \ref{fig:statistics_2}, the calibration-based uncertainty estimation approach is more commonly leveraged to improve efficiency. Its calculation is convenient and sufficient. Accurate confidence information allows the model to improve efficiency in a more optimal way. 

An uncertainty-based early exit in NLP is a technique to improve the efficiency of neural models. As shown in Fig. \ref{fig:early_exit}, early exit allows the model to exit the sequence being processed early based on uncertainty about the predicted output. This strategy works by introducing a threshold value that determines when the model should stop computation in advance via the uncertainty of the predicted output. The threshold value can be set through the desired trade-off between accuracy and efficiency. If the uncertainty score of the prediction is below the threshold, the model continues processing the sequence. Yet if the uncertainty score exceeds the threshold, the model exits early and returns the prediction. This approach can significantly reduce the computation required to process the sequence and improve the overall efficiency of the model.

To name a few, Schuster et al. \cite{schuster2022confident} propose confidence adaptive language modeling to improve the efficiency of autoregressive decoding in PLMs. Confidence-based uncertainty estimation allows PLMs to generate new tokens with intermediate network layers rather than the full model. Thereby it reduces the amount of computation. The approach builds on distribution-free uncertainty quantification and provides a principled approach to improving model efficiency while maintaining predictive quality. Liu et al. \cite{liu-etal-2020-fastbert} design a pre-trained model called FastBERT that applies self-distillation and knowledge distillation to achieve faster and more accurate results in Chinese and English sentence classification tasks. To further improve efficiency, FastBERT utilizes normalized entropy for uncertainty estimation, removing cases with low uncertainty from the batch and sending cases with high uncertainty to the next layer for further reasoning. Xin et al. \cite{xinetal2021berxit} propose BERxiT, which combines \emph{BERT} and \emph{exit} for regression tasks in NLP. Furthermore, Table \ref{tab:app_systems} shows the application is mainly general performance metrics and efficiency improvement comparison metrics. By using uncertainty-based early exit, NLP models can achieve faster processing times and reduce computational costs without sacrificing accuracy.

Overall, uncertainty estimation has the potential to enhance the performance and efficiency of NLP systems. However, one challenge is how to incorporate uncertainty estimates into the decision-making process of NLP systems without introducing bias or overconfidence. Additionally, it is essential to trade off the computational cost of uncertainty estimation with its potential benefits in terms of performance and efficiency.

\subsubsection{Uncertainty and Performance}
Uncertainty information can also be used to guide the model and improve its performance. For instance, Xiao et al. \cite{xiao2019uncerNLP} show that applying uncertainty estimation to NLP tasks such as sentiment analysis, NER, and language modeling models can learn accurate mappings. 
Additionally, uncertainty can provide richer embedding information. By incorporating uncertainty into the model, knowledge graph embeddings can improve the accuracy and robustness of models. Uncertain-aware embeddings can incorporate probabilistic uncertainty into the embedding process, which allows for more precise representations \cite{chen2019embedding,kertkeidkachorn2020gtranse,boutouhami2020uncertain,chen2021probabilistic}. It can be seen from Table \ref{tab:app_systems} that, in addition to performance indicators, uncertainty estimation is mainly exploited for uncertainty calibration indicators.

Uncertainty can also be incorporated into the model's training and inference stages to improve model performance. Gui et al. \cite{guietal2020uncertainty} propose to improve NER task performers with MCD, which predicts error-prone draft labels in the first stage. A two-stream self-attention model in the second stage aims to refine these draft predictions. This approach helps to prevent error propagation and improves the accuracy of NLP models. In addition, Zhou et al. \cite{zhou2020uncertainty} assume that language features alone cannot completely solve the difficulties encountered by the MT model. They propose an adaptive curriculum learning strategy to evaluate the uncertain information of the model. This strategy enhances the performance of neural machine translation, taking into account the complexity and rarity of information in translation pairs. In MT tasks, a source-language sentence may have multiple semantically similar expressions. And a source-language sentence may also have multiple valid counterparts in another language. Thus such tasks are many-to-many problems, which inherently contain uncertainty. Wei et al. \cite{wei2020uncertainty} leverage confidence and KL distributions to capture the relationship between multiple semantically equivalent source sentences and use general semantic information to enhance hidden representations for better translation results. Accurate uncertainty refers to the process of adjusting a model's uncertainty estimate to better reflect its true performance. This helps ensure the model makes reliable predictions and improves its accuracy.

\begin{table}
    \renewcommand{\arraystretch}{1.3}
    \caption{Uncertainty-based applications related to NLP system evaluation.}
    \label{tab:app_evaluation}
    \centering
    \begin{tabular}{lll}
        \toprule
        Tasks Types&UE\_Methods &Metrics\\
        \midrule

        \makecell[l]{MT \cite{zerva2022disentangling,beck2016exploring}\\\cite{glushkova2021uncertainty};
        \\ \\
        \\QA\cite{jiang2021can,lin2022teaching}\\\cite{kadavath2022largeModelKnow,zhou2023navigating}.} &\makecell[l]{Confidence\cite{jiang2021can},\\ \cite{lin2022teaching,kadavath2022largeModelKnow,glushkova2021uncertainty};\\ \\TS\cite{jiang2021can,kadavath2022largeModelKnow};\\ \\GPs\cite{beck2016exploring};\\ \\
        MCD\cite{glushkova2021uncertainty,zerva2022disentangling};\\  \\Ensemble\cite{glushkova2021uncertainty};\\\\ Heteroscedastic\cite{zerva2022disentangling}.}&\makecell[l]{Accuracy\cite{kadavath2022largeModelKnow}\\\cite{jiang2021can,lin2022teaching};\\
        MSE\cite{lin2022teaching};\\NLPD\cite{beck2016exploring};\\
        NLL\cite{beck2016exploring,glushkova2021uncertainty}\\\cite{zerva2022disentangling};\\MAE\cite{beck2016exploring};\\
        ECE\cite{jiang2021can,glushkova2021uncertainty}\\\cite{kadavath2022largeModelKnow,zerva2022disentangling};\\ AUROC\cite{kadavath2022largeModelKnow}; \\BS\cite{kadavath2022largeModelKnow};\\
        Sharpness\cite{glushkova2021uncertainty,zerva2022disentangling};\\ 
        Pearson \cite{glushkova2021uncertainty,zerva2022disentangling}.}\\

        \makecell[l]{Document Quality\\ Assessment\cite{shen2019modelling}.} &GPs\cite{shen2019modelling}. &\makecell[l]{NLPD\cite{shen2019modelling}; \\ RMSE\cite{shen2019modelling};\\
        Pearson\cite{shen2019modelling}.}\\
        \bottomrule
        \end{tabular}
\end{table}

\subsection{Reliability and Trustworthy Assessment}
In this section, we review the applications of uncertainty in the output stage of NLP system. Here the main insight is that \emph{can we fully trust the output of a neural network?} The answer is obviously \emph{no}. The main reason is that its output relies on probability and thereby mistakes are inevitable. The question can be converted to \emph{how much can we trust the output?} In other words, it means reliability or trustworthy extents. Since uncertainty is reflected as a numerical value, which naturally helps us to assess the quality of the model or output. 

By quantifying and analyzing uncertainty, we can gain valuable insights into the strengths and weaknesses of our NLP models, and better understand the risks and limitations of the decisions we make based on them. 
For instance, if an NLP model has high uncertainty in its predictions for certain data points or scenarios, we expect to investigate further or use additional information to improve the performance of the model. Uncertainty can be used to evaluate models' performance. Confidence scores can be compared to the actual correctness of decisions made by the model. This approach enables us to determine how reliable and trustworthy the model is in terms of making accurate decisions. In particular, some metrics given in Sec \ref{ue_metrics} can be exploited to evaluate the confidence calibration and uncertainty classification ability of the model, and then reflect the advantages and disadvantages of the model. 


Furthermore, assessing the reliability of generated content is crucial. Especially in the exponential search space of the generative model, the use of uncertainty for model and output analysis is valuable. Table \ref{tab:app_evaluation} shows that in recent years, research on uncertainty estimation in MT and QA has gradually increased, and various evaluation indicators have increased to explain the model or output results. Typically, multiple candidate translations can be generated in MT, making it difficult to determine the best translation. Estimating the uncertainty of each candidate translation can help researchers choose the translation with the lowest uncertainty, leading to better translation quality \cite{beck2016exploring,glushkova2021uncertainty,zerva2022disentangling}. Recently, as PLMs have become more powerful and widely used, it has become increasingly important to be able to explain how they arrive at answers and provide some degree of uncertainty or confidence in predictions. There are various techniques for evaluating the performance and credibility of PLMs, including self-evaluation techniques, calibration scores, and other methods \cite{jiang2021can,lin2022teaching,kadavath2022largeModelKnow,zhou2023navigating}. These techniques can help to ensure that NLP systems are trustworthy and reliable, and can be used with confidence by users in a range of different applications.

Last but not least, an intuitive question is \textit{how to express uncertainty to improve model evaluation and obtain valuable information?} This is a natural challenge because language expression itself is inherently uncertain. Quantifying the uncertainty expressed by the model, making effective evaluations, and reducing the interaction bias caused by expression are all important considerations. We provide a more detailed discussion of this challenge in Sec \ref{express_uncer}.

\section{Challenges And Future Directions}\label{Challenges_Directions}
\subsection{Extremely High-Dimensional Language-space}

The field of NLP is challenged by the extremely high-dimensional language space, as highlighted by the need to estimate predictive entropy, which requires taking an expectation in output space. This output space has a dimensionality of $\mathcal{O}(|\mathcal{T}|^{N})$, which presents significant computational challenges. Additionally, the lack of a normalized probability density function over sentences necessitates approximating the expectation using Monte Carlo integration, which involves averaging the likelihoods of a finite set of sampled sentences. However, Monte Carlo integration becomes difficult for entropy as it is often dominated by low-probability sentences that have large and negative logs. MT models, for example, can contain hundreds of millions of parameters, which exponentially increases the search space. Additionally, with only a single reference for each source sentence, it becomes difficult to measure the fitness of MT models to data distributions. As a result, researchers face significant scientific challenges in adapting tools from both the machine learning and statistics fields to effectively tackle this problem.

In addition, the complexity of uncertainty in NLP systems is also constantly increasing due to the evolving models and the demand for human-AI interaction. Although the current uncertainty estimation techniques are rich, further exploration is still needed to address issues such as time cost and inference accuracy, and \textit{how to consider the impact of natural language features on uncertainty estimation}. Some uncertainty estimation techniques can also be limited in their \textit{scalability}. Firstly, TS is a convenient calibration technique, but the performance of TS is very sensitive to the choice of the scaling factor. Choosing the optimal scaling factor may require extensive hyperparameter tuning, which can be time-consuming and computationally expensive. Secondly, ensembling multiple models or MCD can be effective for uncertainty estimation but can also be resource-intensive and difficult to scale to large models. Finally, distribution-based uncertainty estimation can provide valuable insights into the uncertainty of DNNs, but it requires distributional assumptions, such as Gaussian or a mixture of Gaussian distributions, which may not apply in all situations and limit its applicability. In the era of large PLMs such as ChatGPT and GPT-4, researchers are not accessible to the model. Even if a PLM is open-sourced, retraining is often necessary, which can be computationally expensive and time-consuming.

\subsection{Variable Length Generation}
The challenge of variable text length in NLP cannot be ignored. One major obstacle is the significant variation in the length of sentences \cite{murray2018correcting}. As Malinin et al. \cite{malinin2018prior} point out, variable-length text generation is particularly challenging in NLG, where longer sequences tend to have lower joint likelihoods due to the conditional independence of token probabilities. In other words, as the length of a sequence increases, its joint likelihood decreases exponentially, leading to a linear increase in its negative log probability. This means that longer sentences contribute more to entropy, which exacerbates the difficulty of dealing with variable-length texts. Text length normalization is commonly adopted to combat variable text length, which assumes that the uncertainty in a sentence is independent of its length \cite{jean2015montreal,koehn2017sixChallenges,murray2018correcting}. Specifically, for estimating the probability $\mathbf{e}=e_{1:l}$ of generating the output sequence, length normalization divides the score $s(\mathbf{e})$ by the length $l$ of the survival text to get $s'(\mathbf{e})=s(\mathbf{e})/l$. For example, Kuhn et al. \cite{kuhn2023semanticeq} apply length normalization of the log probability when estimating the semantic entropy of a QA system.  However, this technique may not be useful in all cases, especially when dealing with long sentences, as it fails to capture the increased complexity and difficulty of longer text sequences.

\subsection{Expression of Uncertainty in Language Model}\label{express_uncer}
In real life, information is rarely black and white, which is why expressing uncertainty is necessary to support the decision-making process. While highly certain expressions are commonly leveraged, uncertain expressions tell us about the confidence, source, and limitations of the information. In the current research, we find two main forms of uncertainty expression: probabilistic expression and natural expression, as demonstrated in Fig. \ref{fig:prompt}. The former one generally has relevant values available for analysis, but the community has found different results on the calibration of neural models. For example, Desai and Durrett \cite{desai2020calibration} show that pre-trained Transformers are relatively well calibrated, while Wang et al. \cite{wang2020inference} find severe miscalibration in MT. Jiang et al. \cite{jiang2021can} study calibration in generative QA and observe only a weak correlation between the log-likelihood model assigned to their answers and the correctness of the answers. In general, PLMs output a possibility for a given tag sequence, but do not output the entire meaning \cite{kuhn2023semanticeq}. 

On the one hand, whether this uncertainty expression is suitable for generative NLP systems remains questionable. We may need stronger correlation indicators to focus on the uncertainty expression of probability. On the other hand, naturalistic representations of uncertainty cover a wide range of discursive behaviors, such as signaling hesitancy, attributing information, or acknowledging limitations, and such representations are intuitive to humans. As the examples displayed in Fig. \ref{fig:prompt}, QA prompts are able to express uncertainty. Through different prefixes, suffixes, or confidence levels of self-assessment outputs, we can obtain numerical uncertainty representations to further analyze the calibration of the model. 

\begin{figure}
\foreach \n in {Different Templates for Uncertainty Representation}
{\begin{tcolorbox}[adjusted title=\n,colframe=blue!75!cyan,colback=blue!1]
    \textbf{Answer Prefix:}\\
    1. \textit{I’m 90\% sure it’s ...}\\
    2. \textit{I vaguely remember it’s ...}\\
    \textbf{Answer Suffix:}\\
    1. \textit{...But I would need to double-check.}\\
    2. \textit{...With 100\% confidence.}\\
    \textbf{Answer Self-evaluation:}\\
    1. \textit{...With what confidence could you answer this question? and output an answer like 0\%, 10\%, 20\%, ..., 100\%.}
\end{tcolorbox}}
\caption{Illustration of a problem template for uncertainty estimation.}
\label{fig:prompt}
\end{figure}

Kadavath et al. \cite{kadavath2022largeModelKnow} expect to observe large benefits of few-shot evaluation with natural language methods, but instead. Yet no major gains are observed in early QA experiments. Zhou et al. \cite{zhou2023navigating} find that non-deterministic expressions can affect language production, and that changes in these expressions can have a substantial impact on overall accuracy, especially when using high-certainty expressions, including in accuracy and calibration. The authors further hypothesize that this may be due to the usage of hyperbolic or exaggerated language in the training set, where numbers are used non-literally. Then, the model somehow recognizes idiomatic, non-literal usage of these extreme values, resulting in lower performance of the task when hints are introduced.

In summary, effective understanding and expression of uncertainty are crucial for NLP applications to ensure better decision-making. While there is a wealth of literature on methods for estimating uncertainty, there is little understanding of how linguistic uncertainty interacts with natural language. This lack of attention has resulted in a poor understanding of how models interact with natural language, leading to challenges in formulating and evaluating uncertainty estimates.

\subsection{NLP and Social Security}\label{NLP_Security}

It is imperative to analyze uncertainty information in NLP systems when integrating social and ethical demands. Therefore, in the following discussion, we will examine future directions from these three perspectives. As an important part of general AI, NLP systems need to account for \textit{moral uncertainty}, that is, try to explain what decisions should be made, given various details of the circumstances of their decisions, including the choices they face and their theoretical rationale. Newberry et al. \cite{newberry2021parliamentary} discuss the idea of using an automated \emph{``moral parliament''} as a way to mitigate the risk of value erosion in AI systems. By incorporating a variety of values from different stakeholders, the AI could be directed by a simulated group representing different values. This approach would better reflect the moral uncertainty of humans and avoid committing AI to one value system. Perhaps the AI system will become a superior version of the current chatbot, surpassing humans in multiple fields \cite{hendrycks2023natural}. As AI systems become more powerful and have a greater impact on society, there is an increasing need to ensure that they are designed with appropriate ethical and moral considerations, despite uncertainty surrounding implementation details across multiple paradigms. This includes developing AI systems that can incorporate ethical uncertainty and allow for multi-stage scrutiny to ensure they are safe and beneficial for society.

Overall, in addition to the limitations and challenges of existing uncertainty estimation methods, there is a need for more effective methods that can provide interpretable and reliable uncertainty estimates. This is an active area of research, and developing better uncertainty estimation methods can help improve the reliability, robustness, and safety of NLP systems.

\section{Conclusion}\label{Conclusion}
Considering the semantic ambiguity of natural language, uncertainty is an important attribute of text information. For the first time, this survey aims to provide a comprehensive review of uncertainty estimation in the NLP field, including its sources, quantification approaches, and applications. We first depict the backgrounds of uncertainty types and summarize the uncertainty sources of an NLP system by its multiple stages. Then, uncertainty estimation methods are systematically reviewed, with pointing out the technical advantages and application difficulties respectively. Meanwhile, evaluation metrics for uncertainty estimation are also discussed. We further investigate relevant literature on uncertainty-aware applications in the NLP field, dividing it into three directions and providing detailed analyses. Finally, we highlight four challenges in the combination between uncertainty estimation and the development of PLMs. Promising research areas are also explored, including the discussion on the scalability of uncertainty technology, the expression of uncertainty, and AI security.

\ifCLASSOPTIONcaptionsoff
  \newpage
\fi


\begin{thebibliography}{100}
\providecommand{\url}[1]{#1}
\csname url@samestyle\endcsname
\providecommand{\newblock}{\relax}
\providecommand{\bibinfo}[2]{#2}
\providecommand{\BIBentrySTDinterwordspacing}{\spaceskip=0pt\relax}
\providecommand{\BIBentryALTinterwordstretchfactor}{4}
\providecommand{\BIBentryALTinterwordspacing}{\spaceskip=\fontdimen2\font plus
\BIBentryALTinterwordstretchfactor\fontdimen3\font minus
  \fontdimen4\font\relax}
\providecommand{\BIBforeignlanguage}[2]{{%
\expandafter\ifx\csname l@#1\endcsname\relax
\typeout{** WARNING: IEEEtran.bst: No hyphenation pattern has been}%
\typeout{** loaded for the language `#1'. Using the pattern for}%
\typeout{** the default language instead.}%
\else
\language=\csname l@#1\endcsname
\fi
#2}}
\providecommand{\BIBdecl}{\relax}
\BIBdecl

\bibitem{sun2022paradigm}
T.-X. Sun, X.-Y. Liu, X.-P. Qiu, and X.-J. Huang, ``Paradigm shift in natural
  language processing,'' \emph{Machine Intelligence Research}, vol.~19, no.~3,
  pp. 169--183, 2022.

\bibitem{siddhantlipton2018deep}
A.~Siddhant and Z.~C. Lipton, ``Deep {B}ayesian active learning for natural
  language processing: Results of a large-scale empirical study,'' in
  \emph{EMNLP}, 2018, pp. 2904--2909.

\bibitem{xiao2019uncerNLP}
Y.~Xiao and W.~Y. Wang, ``Quantifying uncertainties in natural language
  processing tasks,'' in \emph{AAAI}, 2019, pp. 7322--7329.

\bibitem{kuhn2023semanticeq}
L.~Kuhn, Y.~Gal, and S.~Farquhar, ``Semantic uncertainty: Linguistic
  invariances for uncertainty estimation in natural language generation,'' in
  \emph{ICLR}, 2023.

\bibitem{ethayarajh2020classifier}
K.~Ethayarajh, ``Is your classifier actually biased? measuring fairness under
  uncertainty with bernstein bounds,'' in \emph{ACL}, 2020, pp. 2914--2919.

\bibitem{kivimaki2022uncertainty}
J.~Kivim{\"a}ki \emph{et~al.}, ``Uncertainty estimation with calibrated
  confidence scores,'' 2022.

\bibitem{pei2022transformer}
J.~Pei, C.~Wang, and G.~Szarvas, ``Transformer uncertainty estimation with
  hierarchical stochastic attention,'' in \emph{AAAI}, 2022, pp.
  11\,147--11\,155.

\bibitem{malinin2018prior}
A.~Malinin and M.~Gales, ``Predictive uncertainty estimation via prior
  networks,'' \emph{NeurIPS}, 2018.

\bibitem{kopetzki2021evaluating}
A.-K. Kopetzki, B.~Charpentier, D.~Z{\"u}gner, S.~Giri, and S.~G{\"u}nnemann,
  ``Evaluating robustness of predictive uncertainty estimation: Are
  dirichlet-based models reliable?'' in \emph{ICML}, 2021, pp. 5707--5718.

\bibitem{guo2017calibration}
C.~Guo, G.~Pleiss, Y.~Sun, and K.~Q. Weinberger, ``On calibration of modern
  neural networks,'' in \emph{ICML}, 2017, pp. 1321--1330.

\bibitem{raffel2020T5}
C.~Raffel, N.~Shazeer, A.~Roberts, K.~Lee, S.~Narang, M.~Matena, Y.~Zhou,
  W.~Li, and P.~J. Liu, ``Exploring the limits of transfer learning with a
  unified text-to-text transformer,'' \emph{JMLR}, pp. 5485--5551, 2020.

\bibitem{radford2018improving}
A.~Radford, K.~Narasimhan, T.~Salimans, I.~Sutskever \emph{et~al.}, ``Improving
  language understanding by generative pre-training,'' 2018.

\bibitem{radford2019language}
A.~Radford, J.~Wu, R.~Child, D.~Luan, D.~Amodei, I.~Sutskever \emph{et~al.},
  ``Language models are unsupervised multitask learners.''

\bibitem{brown2020language}
T.~Brown, B.~Mann, N.~Ryder, M.~Subbiah, J.~D. Kaplan, P.~Dhariwal,
  A.~Neelakantan, P.~Shyam, G.~Sastry, A.~Askell \emph{et~al.}, ``Language
  models are few-shot learners,'' \emph{NeurIPS}, pp. 1877--1901, 2020.

\bibitem{xu2020understanding}
J.~Xu, S.~Desai, and G.~Durrett, ``Understanding neural abstractive
  summarization models via uncertainty,'' in \emph{EMNLP}, 2020, pp.
  6275--6281.

\bibitem{gawlikowski2021dnnUncerSurvey}
J.~Gawlikowski, C.~R.~N. Tassi, M.~Ali, J.~Lee, M.~Humt, J.~Feng, A.~Kruspe,
  R.~Triebel, P.~Jung, R.~Roscher \emph{et~al.}, ``A survey of uncertainty in
  deep neural networks,'' \emph{arXiv preprint arXiv:2107.03342}, 2021.

\bibitem{hullermeier2021aleatoric}
E.~H{\"u}llermeier and W.~Waegeman, ``Aleatoric and epistemic uncertainty in
  machine learning: An introduction to concepts and methods,'' \emph{Machine
  Learning}, pp. 457--506, 2021.

\bibitem{zerva2022disentangling}
C.~Zerva, T.~Glushkova, R.~Rei, and A.~F. Martins, ``Disentangling uncertainty
  in machine translation evaluation,'' in \emph{EMNLP}, 2022, pp. 8622--8641.

\bibitem{delaforge2022ebbe}
A.~Delaforge, J.~Az{\'e}, S.~Bringay, C.~Mollevi, A.~Sallaberry, and
  M.~Servajean, ``Ebbe-text: Explaining neural networks by exploring text
  classification decision boundaries,'' \emph{IEEE Transactions on
  Visualization and Computer Graphics}, 2022.

\bibitem{dragos2013ontological}
V.~Dragos, ``An ontological analysis of uncertainty in soft data,'' in
  \emph{FUSION}, 2013, pp. 1566--1573.

\bibitem{blodgett2020Inconsistencies}
S.~L. Blodgett, S.~Barocas, H.~Daum{\'e}~III, and H.~Wallach, ``Language
  (technology) is power: A critical survey of {``}bias{''} in {NLP},'' in
  \emph{ACL}, 2020, pp. 5454--5476.

\bibitem{Venuti2008Translator}
L.~VENUTI, ``The translator's invisibility,'' \emph{Criticism}, pp. 179--212,
  1986.

\bibitem{ott2018MTuncer}
M.~Ott, M.~Auli, D.~Grangier, and M.~Ranzato, ``Analyzing uncertainty in neural
  machine translation,'' in \emph{ICML}, 2018, pp. 3956--3965.

\bibitem{levy-2008-noisy}
R.~Levy, ``A noisy-channel model of human sentence comprehension under
  uncertain input,'' in \emph{EMNLP}, 2008, pp. 234--243.

\bibitem{liu-etal-2018-word_noise}
T.~Liu, X.~Zhang, W.~Zhou, and W.~Jia, ``Neural relation extraction via
  inner-sentence noise reduction and transfer learning,'' in \emph{EMNLP},
  2018, pp. 2195--2204.

\bibitem{kadavath2022largeModelKnow}
S.~Kadavath, T.~Conerly, A.~Askell, T.~Henighan, D.~Drain, E.~Perez,
  N.~Schiefer, Z.~H. Dodds, N.~DasSarma, E.~Tran-Johnson \emph{et~al.},
  ``Language models (mostly) know what they know,'' \emph{arXiv preprint
  arXiv:2207.05221}, 2022.

\bibitem{wei2020uncertainty}
X.~Wei, H.~Yu, Y.~Hu, R.~Weng, L.~Xing, and W.~Luo, ``Uncertainty-aware
  semantic augmentation for neural machine translation,'' in \emph{EMNLP},
  2020, pp. 2724--2735.

\bibitem{qin2022sun}
B.~Qin, L.~Wang, B.~Hui, B.~Li, X.~Wei, B.~Li, F.~Huang, L.~Si, M.~Yang, and
  Y.~Li, ``{SUN}: Exploring intrinsic uncertainties in text-to-{SQL} parsers,''
  in \emph{COLING}, 2022, pp. 5298--5308.

\bibitem{arora-etal-2021-types}
U.~Arora, W.~Huang, and H.~He, ``Types of out-of-distribution texts and how to
  detect them,'' in \emph{EMNLP}, 2021, pp. 10\,687--10\,701.

\bibitem{kamath-etal-2020-selective}
A.~Kamath, R.~Jia, and P.~Liang, ``Selective question answering under domain
  shift,'' in \emph{ACL}, 2020, pp. 5684--5696.

\bibitem{carlebach-etal-2020-news}
M.~Carlebach, R.~Cheruvu, B.~Walker, C.~Ilharco~Magalhaes, and S.~Jaume, ``News
  aggregation with diverse viewpoint identification using neural embeddings and
  semantic understanding models,'' in \emph{Argument Mining}, 2020, pp. 59--66.

\bibitem{xinetal2021select2art}
J.~Xin, R.~Tang, Y.~Yu, and J.~Lin, ``The art of abstention: Selective
  prediction and error regularization for natural language processing,'' in
  \emph{ACL-IJCNLP}, 2021, pp. 1040--1051.

\bibitem{battaglia2018relational}
P.~W. Battaglia, J.~B. Hamrick, V.~Bapst, A.~Sanchez-Gonzalez, V.~Zambaldi,
  M.~Malinowski, A.~Tacchetti, D.~Raposo, A.~Santoro, R.~Faulkner
  \emph{et~al.}, ``Relational inductive biases, deep learning, and graph
  networks,'' \emph{arXiv preprint arXiv:1806.01261}, 2018.

\bibitem{lakshminarayanan2017simple}
B.~Lakshminarayanan, A.~Pritzel, and C.~Blundell, ``Simple and scalable
  predictive uncertainty estimation using deep ensembles,'' \emph{NeurIPS},
  2017.

\bibitem{amini2020deep}
A.~Amini, W.~Schwarting, A.~Soleimany, and D.~Rus, ``Deep evidential
  regression,'' \emph{NeurIPS}, pp. 14\,927--14\,937, 2020.

\bibitem{liu-etal-2020-fastbert}
W.~Liu, P.~Zhou, Z.~Wang, Z.~Zhao, H.~Deng, and Q.~Ju, ``{F}ast{BERT}: a
  self-distilling {BERT} with adaptive inference time,'' in \emph{ACL}, 2020,
  pp. 6035--6044.

\bibitem{guietal2020uncertainty}
T.~Gui, J.~Ye, Q.~Zhang, Z.~Li, Z.~Fei, Y.~Gong, and X.~Huang,
  ``Uncertainty-aware label refinement for sequence labeling,'' in
  \emph{EMNLP}, 2020, pp. 2316--2326.

\bibitem{schuster2022confident}
T.~Schuster, A.~Fisch, J.~Gupta, M.~Dehghani, D.~Bahri, V.~Tran, Y.~Tay, and
  D.~Metzler, ``Confident adaptive language modeling,'' \emph{NeurIPS}, pp.
  17\,456--17\,472, 2022.

\bibitem{fedus2022switch}
W.~Fedus, B.~Zoph, and N.~Shazeer, ``Switch transformers: Scaling to trillion
  parameter models with simple and efficient sparsity,'' \emph{The Journal of
  Machine Learning Research}, vol.~23, no.~1, pp. 5232--5270, 2022.

\bibitem{beck2020representation}
C.~Beck, H.~Booth, M.~El-Assady, and M.~Butt, ``Representation problems in
  linguistic annotations: Ambiguity, variation, uncertainty, error and bias,''
  in \emph{14th Linguistic Annotation Workshop}, 2020, pp. 60--73.

\bibitem{bonneau2014overview}
G.-P. Bonneau, H.-C. Hege, C.~R. Johnson, M.~M. Oliveira, K.~Potter,
  P.~Rheingans, and T.~Schultz, ``Overview and state-of-the-art of uncertainty
  visualization,'' \emph{Scientific Visualization: Uncertainty, Multifield,
  Biomedical, and Scalable Visualization}, pp. 3--27, 2014.

\bibitem{John2017UncertaintyIV}
M.~John, ``Uncertainty in visual text analysis in the context of the digital
  humanities,'' 2017.

\bibitem{bender-friedman-2018-data}
E.~M. Bender and B.~Friedman, ``Data statements for natural language
  processing: Toward mitigating system bias and enabling better science,''
  \emph{TACL}, pp. 587--604, 2018.

\bibitem{dodge2020fine}
J.~Dodge, G.~Ilharco, R.~Schwartz, A.~Farhadi, H.~Hajishirzi, and N.~Smith,
  ``Fine-tuning pretrained language models: Weight initializations, data
  orders, and early stopping,'' \emph{arXiv preprint arXiv:2002.06305}, 2020.

\bibitem{OpenNER}
C.~Liang, Y.~Yu, H.~Jiang, S.~Er, R.~Wang, T.~Zhao, and C.~Zhang, ``Bond:
  Bert-assisted open-domain named entity recognition with distant
  supervision,'' 2020.

\bibitem{pandey2022multilinguals}
A.~Pandey, S.~Daw, N.~Unnam, and V.~Pudi, ``Multilinguals at {S}em{E}val-2022
  task 11: Complex {NER} in semantically ambiguous settings for low resource
  languages,'' in \emph{SemEval}, 2022, pp. 1469--1476.

\bibitem{panchendrarajan2018crfner}
R.~Panchendrarajan and A.~Amaresan, ``Bidirectional lstm-crf for named entity
  recognition,'' in \emph{Proceedings of the 32nd Pacific Asia conference on
  language, information and computation}, 2018.

\bibitem{Li2022NERsurvey}
J.~Li, A.~Sun, J.~Han, and C.~Li, ``A survey on deep learning for named entity
  recognition,'' \emph{IEEE Trans. on Knowl. and Data Eng.}, p. 50–70, 2022.

\bibitem{zhai2017neural}
F.~Zhai, S.~Potdar, B.~Xiang, and B.~Zhou, ``Neural models for sequence
  chunking,'' in \emph{AAAI}, 2017.

\bibitem{partalas-etal-2016-learning}
I.~Partalas, C.~Lopez, N.~Derbas, and R.~Kalitvianski, ``Learning to search for
  recognizing named entities in {T}witter,'' in \emph{WNUT}, 2016, pp.
  171--177.

\bibitem{levygoldberg2014linguistic}
O.~Levy and Y.~Goldberg, ``Linguistic regularities in sparse and explicit word
  representations,'' in \emph{CoNLL}, 2014, pp. 171--180.

\bibitem{kuleshov2015calibrated}
V.~Kuleshov and P.~S. Liang, ``Calibrated structured prediction,''
  \emph{NeurIPS}, 2015.

\bibitem{stent2005evaluating}
A.~Stent, M.~Marge, and M.~Singhai, ``Evaluating evaluation methods for
  generation in the presence of variation.'' in \emph{CICLing}, 2005, pp.
  341--351.

\bibitem{bai2022training}
Y.~Bai, A.~Jones, K.~Ndousse, A.~Askell, A.~Chen, N.~DasSarma, D.~Drain,
  S.~Fort, D.~Ganguli, T.~Henighan \emph{et~al.}, ``Training a helpful and
  harmless assistant with reinforcement learning from human feedback,''
  \emph{arXiv preprint arXiv:2204.05862}, 2022.

\bibitem{murray2018correcting}
K.~Murray and D.~Chiang, ``Correcting length bias in neural machine
  translation,'' in \emph{Proceedings of the Third Conference on Machine
  Translation: Research Papers}, 2018, pp. 212--223.

\bibitem{malininuncertainty}
A.~Malinin and M.~Gales, ``Uncertainty estimation in autoregressive structured
  prediction,'' in \emph{ICLR}, 2022.

\bibitem{mikolov2013distributed}
T.~Mikolov, I.~Sutskever, K.~Chen, G.~S. Corrado, and J.~Dean, ``Distributed
  representations of words and phrases and their compositionality,''
  \emph{NeurIPS}, 2013.

\bibitem{wang-etal-2022-uncertainty}
\BIBentryALTinterwordspacing
Y.~Wang, D.~Beck, T.~Baldwin, and K.~Verspoor, ``Uncertainty estimation and
  reduction of pre-trained models for text regression,'' \emph{Transactions of
  the Association for Computational Linguistics}, vol.~10, pp. 680--696, 2022.
  [Online]. Available: \url{https://aclanthology.org/2022.tacl-1.39}
\BIBentrySTDinterwordspacing

\bibitem{guzman-etal-2019-flores}
\BIBentryALTinterwordspacing
F.~Guzm{\'a}n, P.-J. Chen, M.~Ott, J.~Pino, G.~Lample, P.~Koehn, V.~Chaudhary,
  and M.~Ranzato, ``The {FLORES} evaluation datasets for low-resource machine
  translation: {N}epali{--}{E}nglish and {S}inhala{--}{E}nglish,'' in
  \emph{EMNLP-IJCNLP}, Hong Kong, China, Nov. 2019, pp. 6098--6111. [Online].
  Available: \url{https://aclanthology.org/D19-1632}
\BIBentrySTDinterwordspacing

\bibitem{corley-mihalcea-2005-measuring}
\BIBentryALTinterwordspacing
C.~Corley and R.~Mihalcea, ``Measuring the semantic similarity of texts,'' in
  \emph{Proceedings of the {ACL} Workshop on Empirical Modeling of Semantic
  Equivalence and Entailment}.\hskip 1em plus 0.5em minus 0.4em\relax Ann
  Arbor, Michigan: Association for Computational Linguistics, Jun. 2005, pp.
  13--18. [Online]. Available: \url{https://aclanthology.org/W05-1203}
\BIBentrySTDinterwordspacing

\bibitem{kuleshov2018accurate}
V.~Kuleshov, N.~Fenner, and S.~Ermon, ``Accurate uncertainties for deep
  learning using calibrated regression,'' in \emph{International conference on
  machine learning}.\hskip 1em plus 0.5em minus 0.4em\relax PMLR, 2018, pp.
  2796--2804.

\bibitem{blundell2015weight}
C.~Blundell, J.~Cornebise, K.~Kavukcuoglu, and D.~Wierstra, ``Weight
  uncertainty in neural network,'' in \emph{ICML}, 2015, pp. 1613--1622.

\bibitem{kingma2015variational}
D.~P. Kingma, T.~Salimans, and M.~Welling, ``Variational dropout and the local
  reparameterization trick,'' \emph{NeurIPS}, 2015.

\bibitem{louizos2016structured}
C.~Louizos and M.~Welling, ``Structured and efficient variational deep learning
  with matrix gaussian posteriors,'' in \emph{ICML}, 2016, pp. 1708--1716.

\bibitem{ruder-plank-2017-learning}
S.~Ruder and B.~Plank, ``Learning to select data for transfer learning with
  {B}ayesian optimization,'' in \emph{EMNLP}, Sep. 2017, pp. 372--382.

\bibitem{hinton2015distilling}
G.~Hinton, O.~Vinyals, and J.~Dean, ``Distilling the knowledge in a neural
  network,'' \emph{arXiv preprint arXiv:1503.02531}, 2015.

\bibitem{mukhoti2020calibrating}
J.~Mukhoti, V.~Kulharia, A.~Sanyal, S.~Golodetz, P.~Torr, and P.~Dokania,
  ``Calibrating deep neural networks using focal loss,'' \emph{NeurIPS}, pp.
  15\,288--15\,299, 2020.

\bibitem{zhu2022boundary}
E.~Zhu and J.~Li, ``Boundary smoothing for named entity recognition,'' in
  \emph{ACL}, 2022, pp. 7096--7108.

\bibitem{xiao-etal-2022-uncertainty}
Y.~Xiao, P.~P. Liang, U.~Bhatt, W.~Neiswanger, R.~Salakhutdinov, and L.-P.
  Morency, ``Uncertainty quantification with pre-trained language models: A
  large-scale empirical analysis,'' in \emph{EMNLP Findings}, Dec. 2022, pp.
  7273--7284.

\bibitem{lin2017focal}
T.-Y. Lin, P.~Goyal, R.~Girshick, K.~He, and P.~Doll{\'a}r, ``Focal loss for
  dense object detection,'' in \emph{Proceedings of the IEEE international
  conference on computer vision}, 2017, pp. 2980--2988.

\bibitem{vovk2005algorithmic}
V.~Vovk, A.~Gammerman, and G.~Shafer, \emph{Algorithmic learning in a random
  world}.

\bibitem{neal1992bayesian}
R.~M. Neal, ``Bayesian training of backpropagation networks by the hybrid monte
  carlo method,'' Citeseer, Tech. Rep., 1992.

\bibitem{gal2016dropout}
Y.~Gal and Z.~Ghahramani, ``Dropout as a bayesian approximation: Representing
  model uncertainty in deep learning,'' in \emph{ICML}, 2016, pp. 1050--1059.

\bibitem{hinton1993keeping}
G.~E. Hinton and D.~Van~Camp, ``Keeping the neural networks simple by
  minimizing the description length of the weights,'' in \emph{6COLT93}, 1993,
  pp. 5--13.

\bibitem{poole2019variational}
B.~Poole, S.~Ozair, A.~Van Den~Oord, A.~Alemi, and G.~Tucker, ``On variational
  bounds of mutual information,'' in \emph{ICML}, 2019, pp. 5171--5180.

\bibitem{yu2023learning}
Y.~Yu, H.~Sajjad, and J.~Xu, ``Learning uncertainty for unknown domains with
  zero-target-assumption,'' in \emph{ICLR}, 2023.

\bibitem{houlsby2011bayesian}
N.~Houlsby, F.~Husz{\'a}r, Z.~Ghahramani, and M.~Lengyel, ``Bayesian active
  learning for classification and preference learning,'' \emph{arXiv preprint
  arXiv:1112.5745}, 2011.

\bibitem{li2021multitaskdense}
M.~Li, M.~Li, K.~Xiong, and J.~Lin, ``Multi-task dense retrieval via model
  uncertainty fusion for open-domain question answering,'' in
  \emph{EMNLP-Findings}, 2021, pp. 274--287.

\bibitem{rainagales2022answer}
V.~Raina and M.~Gales, ``Answer uncertainty and unanswerability in
  multiple-choice machine reading comprehension,'' in \emph{ACL-Findings},
  2022, pp. 1020--1034.

\bibitem{andersenmaalej2022efficient}
J.~S. Andersen and W.~Maalej, ``Efficient, uncertainty-based moderation of
  neural networks text classifiers,'' in \emph{ACL-Findings}, 2022, pp.
  1536--1546.

\bibitem{pop2018deep}
R.~Pop and P.~Fulop, ``Deep ensemble bayesian active learning: Addressing the
  mode collapse issue in monte carlo dropout via ensembles,'' \emph{arXiv
  preprint arXiv:1811.03897}, 2018.

\bibitem{ovadia2019can}
Y.~Ovadia, E.~Fertig, J.~Ren, Z.~Nado, D.~Sculley, S.~Nowozin, J.~Dillon,
  B.~Lakshminarayanan, and J.~Snoek, ``Can you trust your model's uncertainty?
  evaluating predictive uncertainty under dataset shift,'' \emph{NeurIPS},
  2019.

\bibitem{fort2019deep}
S.~Fort, H.~Hu, and B.~Lakshminarayanan, ``Deep ensembles: A loss landscape
  perspective,'' \emph{arXiv preprint arXiv:1912.02757}, 2019.

\bibitem{sensoy2018evidential}
M.~Sensoy, L.~Kaplan, and M.~Kandemir, ``Evidential deep learning to quantify
  classification uncertainty,'' \emph{NeurIPS}, 2018.

\bibitem{charpentier2020posterior}
B.~Charpentier, D.~Z{\"u}gner, and S.~G{\"u}nnemann, ``Posterior network:
  Uncertainty estimation without ood samples via density-based pseudo-counts,''
  \emph{NeurIPS}, pp. 1356--1367, 2020.

\bibitem{zhang2023ner}
Z.~Zhang, M.~Hu, S.~Zhao, M.~Huang, H.~Wang, L.~Liu, Z.~Zhang, Z.~Liu, and
  B.~Wu, ``E-ner: Evidential deep learning for trustworthy named entity
  recognition,'' \emph{arXiv preprint arXiv:2305.17854}, 2023.

\bibitem{vilnis2014word}
L.~Vilnis and A.~McCallum, ``Word representations via gaussian embedding,''
  \emph{arXiv preprint arXiv:1412.6623}, 2014.

\bibitem{zhouetal2021contrastive}
W.~Zhou, F.~Liu, and M.~Chen, ``Contrastive out-of-distribution detection for
  pretrained transformers,'' in \emph{EMNLP}, 2021, pp. 1100--1111.

\bibitem{vazhentsevetal2022uncertainty}
A.~Vazhentsev, G.~Kuzmin, A.~Shelmanov, A.~Tsvigun, E.~Tsymbalov, K.~Fedyanin,
  M.~Panov, A.~Panchenko, G.~Gusev, M.~Burtsev, M.~Avetisian, and L.~Zhukov,
  ``Uncertainty estimation of transformer predictions for misclassification
  detection,'' in \emph{ACL}, 2022, pp. 8237--8252.

\bibitem{williams2006gaussian}
C.~K. Williams and C.~E. Rasmussen, \emph{Gaussian processes for machine
  learning}, 2006.

\bibitem{shah2013investigation}
K.~Shah, T.~Conn, and L.~Specia, ``An investigation on the effectiveness of
  features for translation quality estimation,'' in \emph{MTSummit}, 2013.

\bibitem{beck2014joint}
D.~Beck, T.~Cohn, and L.~Specia, ``Joint emotion analysis via multi-task
  gaussian processes,'' in \emph{EMNLP}, 2014, pp. 1798--1803.

\bibitem{beck2016exploring}
D.~Beck, L.~Specia, and T.~Cohn, ``Exploring prediction uncertainty in machine
  translation quality estimation,'' in \emph{SIGNLL}, 2016, pp. 208--218.

\bibitem{assel2017brier}
M.~Assel, D.~D. Sjoberg, and A.~J. Vickers, ``The brier score does not evaluate
  the clinical utility of diagnostic tests or prediction models,''
  \emph{Diagnostic and prognostic research}, pp. 1--7, 2017.

\bibitem{quinonero2006evaluating}
J.~Quinonero-Candela, C.~E. Rasmussen, F.~Sinz, O.~Bousquet, and B.~Scholkopf,
  ``Evaluating predictive uncertainty challenge,'' \emph{Lecture Notes in
  Computer Science}, pp. 1--27, 2006.

\bibitem{lin2023on}
Z.~Lin, D.~Phan, P.~Pasupat, J.~Z. Liu, and J.~Shang, ``On compositional
  uncertainty quantification for seq2seq graph parsing,'' in \emph{ICLR}, 2023.

\bibitem{Yaniv2010Selective}
R.~El-Yaniv and Y.~Wiener, ``On the foundations of noise-free selective
  classification,'' \emph{J. Mach. Learn. Res.}, p. 1605–1641, 2010.

\bibitem{xinetal2021art}
J.~Xin, R.~Tang, Y.~Yu, and J.~Lin, ``The art of abstention: Selective
  prediction and error regularization for natural language processing,'' in
  \emph{ACL-IJCNLP}, 2021, pp. 1040--1051.

\bibitem{zhu2008active}
J.~Zhu, H.~Wang, T.~Yao, and B.~K. Tsou, ``Active learning with sampling by
  uncertainty and density for word sense disambiguation and text
  classification,'' in \emph{COLING}, 2008, pp. 1137--1144.

\bibitem{yuanetal2020cold}
M.~Yuan, H.-T. Lin, and J.~Boyd-Graber, ``Cold-start active learning through
  self-supervised language modeling,'' in \emph{EMNLP}, 2020, pp. 7935--7948.

\bibitem{eindor2020active}
L.~Ein-Dor, A.~Halfon, A.~Gera, E.~Shnarch, L.~Dankin, L.~Choshen,
  M.~Danilevsky, R.~Aharonov, Y.~Katz, and N.~Slonim, ``{A}ctive {L}earning for
  {BERT}: {A}n {E}mpirical {S}tudy,'' in \emph{EMNLP}, 2020, pp. 7949--7962.

\bibitem{yu2022actune}
Y.~Yu, L.~Kong, J.~Zhang, R.~Zhang, and C.~Zhang, ``{A}c{T}une:
  Uncertainty-based active self-training for active fine-tuning of pretrained
  language models,'' in \emph{NAACL}, 2022, pp. 1422--1436.

\bibitem{margatina2021active}
K.~Margatina, G.~Vernikos, L.~Barrault, and N.~Aletras, ``Active learning by
  acquiring contrastive examples,'' in \emph{EMNLP}, 2021, pp. 650--663.

\bibitem{ru2020active}
D.~Ru, J.~Feng, L.~Qiu, H.~Zhou, M.~Wang, W.~Zhang, Y.~Yu, and L.~Li, ``Active
  sentence learning by adversarial uncertainty sampling in discrete space,'' in
  \emph{EMNLP}, 2020, pp. 4908--4917.

\bibitem{NEURIPS2020_f23d125d}
S.~Mukherjee and A.~Awadallah, ``Uncertainty-aware self-training for few-shot
  text classification,'' in \emph{NeurIPS}, H.~Larochelle, M.~Ranzato,
  R.~Hadsell, M.~Balcan, and H.~Lin, Eds., 2020, pp. 21\,199--21\,212.

\bibitem{lei2022uncertainty}
S.~Lei, X.~Zhang, J.~He, F.~Chen, and C.-T. Lu, ``Uncertainty-aware
  cross-lingual transfer with pseudo partial labels,'' in
  \emph{NAACL-Findings}, 2022, pp. 1987--1997.

\bibitem{shelmanov2021active}
A.~Shelmanov, D.~Puzyrev, L.~Kupriyanova, D.~Belyakov, D.~Larionov, N.~Khromov,
  O.~Kozlova, E.~Artemova, D.~V. Dylov, and A.~Panchenko, ``Active learning for
  sequence tagging with deep pre-trained models and {B}ayesian uncertainty
  estimates,'' in \emph{EACL}, 2021, pp. 1698--1712.

\bibitem{chaudhary2019little}
A.~Chaudhary, J.~Xie, Z.~Sheikh, G.~Neubig, and J.~Carbonell, ``A little
  annotation does a lot of good: A study in bootstrapping low-resource named
  entity recognizers,'' in \emph{EMNLP-IJCNLP}, 2019, pp. 5164--5174.

\bibitem{liu2022ltp}
M.~Liu, Z.~Tu, T.~Zhang, T.~Su, X.~Xu, and Z.~Wang, ``Ltp: a new active
  learning strategy for crf-based named entity recognition,'' \emph{Neural
  Processing Letters}, pp. 2433--2454, 2022.

\bibitem{lyu2020you}
Z.~Lyu, D.~Duolikun, B.~Dai, Y.~Yao, P.~Minervini, T.~Z. Xiao, and Y.~Gal,
  ``You need only uncertain answers: Data efficient multilingual question
  answering,'' \emph{TWorkshop on Uncertainty and Ro-Bustness in Deep
  Learning}, 2020.

\bibitem{gidiotis2022trust}
A.~Gidiotis and G.~Tsoumakas, ``Should we trust this summary? {B}ayesian
  abstractive summarization to the rescue,'' in \emph{ACL-Findings}, 2022, pp.
  4119--4131.

\bibitem{hendrycks2020pretrained}
D.~Hendrycks, X.~Liu, E.~Wallace, A.~Dziedzic, R.~Krishnan, and D.~Song,
  ``Pretrained transformers improve out-of-distribution robustness,'' in
  \emph{ACL}, 2020, pp. 2744--2751.

\bibitem{shen2021enhancing}
Y.~Shen, Y.-C. Hsu, A.~Ray, and H.~Jin, ``Enhancing the generalization for
  intent classification and out-of-domain detection in slu,'' in \emph{COLING},
  2021, pp. 2443--2453.

\bibitem{desai2020calibration}
S.~Desai and G.~Durrett, ``Calibration of pre-trained transformers,'' in
  \emph{EMNLP}, 2020, pp. 295--302.

\bibitem{zhang2019mitigating}
X.~Zhang, F.~Chen, C.-T. Lu, and N.~Ramakrishnan, ``Mitigating uncertainty in
  document classification,'' in \emph{NAACL}, 2019, pp. 3126--3136.

\bibitem{he2020towards}
J.~He, X.~Zhang, S.~Lei, Z.~Chen, F.~Chen, A.~Alhamadani, B.~Xiao, and C.~Lu,
  ``Towards more accurate uncertainty estimation in text classification,'' in
  \emph{EMNLP}, 2020, pp. 8362--8372.

\bibitem{Hu2021URTX}
Y.~Hu and L.~Khan, ``Uncertainty-aware reliable text classification,'' in
  \emph{SIGKDD}, 2021, p. 628–636.

\bibitem{xiao2020wat}
T.~Z. Xiao, A.~N. Gomez, and Y.~Gal, ``Wat zei je? detecting
  out-of-distribution translations with variational transformers,'' \emph{arXiv
  preprint arXiv:2006.08344}, 2020.

\bibitem{wu2021uncertainty}
M.~Wu, Y.~Li, M.~Zhang, L.~Li, G.~Haffari, and Q.~Liu, ``Uncertainty-aware
  balancing for multilingual and multi-domain neural machine translation
  training,'' in \emph{EMNLP}, 2021, pp. 7291--7305.

\bibitem{gargmoschitti2021will}
S.~Garg and A.~Moschitti, ``Will this question be answered? question filtering
  via answer model distillation for efficient question answering,'' in
  \emph{EMNLP}, 2021, pp. 7329--7346.

\bibitem{varshneyetal2022investigating}
N.~Varshney, S.~Mishra, and C.~Baral, ``Investigating selective prediction
  approaches across several tasks in {IID}, {OOD}, and adversarial settings,''
  in \emph{ACL-Findings}, 2022, pp. 1995--2002.

\bibitem{varshney2022towards}
N.~\vspace{0mm}Varshney, S.~Mishra, and C.~Baral, ``Towards improving selective
  prediction ability of nlp systems,'' in \emph{RepL4NLP}, 2022, pp. 221--226.

\bibitem{hendrycks2017a}
D.~Hendrycks and K.~Gimpel, ``A baseline for detecting misclassified and
  out-of-distribution examples in neural networks,'' in \emph{ICLR}, 2017.

\bibitem{hendrycks2019deep}
D.~Hendrycks, M.~Mazeika, and T.~Dietterich, ``Deep anomaly detection with
  outlier exposure,'' in \emph{ICLR}, 2019.

\bibitem{Zhang2019Reply}
Q.~Zhang, A.~Lipani, S.~Liang, and E.~Yilmaz, ``Reply-aided detection of
  misinformation via bayesian deep learning,'' in \emph{WWW}, 2019, p.
  2333–2343.

\bibitem{kochkinaliakata2020estimating}
E.~Kochkina and M.~Liakata, ``Estimating predictive uncertainty for rumour
  verification models,'' in \emph{ACL}, 2020, pp. 6964--6981.

\bibitem{fengetal2020none}
Y.~Feng, S.~Mehri, M.~Eskenazi, and T.~Zhao, ``{``}none of the above{''}:
  Measure uncertainty in dialog response retrieval,'' in \emph{ACL}, 2020, pp.
  2013--2020.

\bibitem{mukhoti2021deterministic}
J.~Mukhoti, A.~Kirsch, J.~van Amersfoort, P.~H. Torr, and Y.~Gal,
  ``Deterministic neural networks with inductive biases capture epistemic and
  aleatoric uncertainty,'' \emph{arXiv preprint arXiv:2102.11582}, p.~13, 2021.

\bibitem{RYU201726}
S.~Ryu, S.~Kim, J.~Choi, H.~Yu, and G.~G. Lee, ``Neural sentence embedding
  using only in-domain sentences for out-of-domain sentence detection in dialog
  systems,'' \emph{Pattern Recognition Letters}, pp. 26--32, 2017.

\bibitem{ryu2018domain}
S.~Ryu, S.~Koo, H.~Yu, and G.~G. Lee, ``Out-of-domain detection based on
  generative adversarial network,'' in \emph{EMNLP}, 2018, pp. 714--718.

\bibitem{goodfellow2014generative}
I.~J. Goodfellow, J.~Pouget-Abadie, M.~Mirza, B.~Xu, D.~Warde-Farley, S.~Ozair,
  A.~Courville, and Y.~Bengio, ``Generative adversarial nets,'' \emph{stat},
  p.~10, 2014.

\bibitem{tan2019domain}
M.~Tan, Y.~Yu, H.~Wang, D.~Wang, S.~Potdar, S.~Chang, and M.~Yu,
  ``Out-of-domain detection for low-resource text classification tasks,'' in
  \emph{EMNLP-IJCNLP}, 2019, pp. 3566--3572.

\bibitem{lee2018training}
K.~Lee, H.~Lee, K.~Lee, and J.~Shin, ``Training confidence-calibrated
  classifiers for detecting out-of-distribution samples,'' in \emph{ICLR},
  2018.

\bibitem{geng2021romebert}
S.~Geng, P.~Gao, Z.~Fu, and Y.~Zhang, ``Romebert: Robust training of multi-exit
  bert,'' \emph{arXiv preprint arXiv:2101.09755}, 2021.

\bibitem{xinetal2021berxit}
J.~Xin, R.~Tang, Y.~Yu, and J.~Lin, ``{BER}xi{T}: Early exiting for {BERT} with
  better fine-tuning and extension to regression,'' in \emph{EACL}, 2021, pp.
  91--104.

\bibitem{zhou2020bert}
W.~Zhou, C.~Xu, T.~Ge, J.~McAuley, K.~Xu, and F.~Wei, ``Bert loses patience:
  Fast and robust inference with early exit,'' \emph{NeurIPS}, pp.
  18\,330--18\,341, 2020.

\bibitem{schwartzetal2020right}
R.~Schwartz, G.~Stanovsky, S.~Swayamdipta, J.~Dodge, and N.~A. Smith, ``The
  right tool for the job: Matching model and instance complexities,'' in
  \emph{ACL}, 2020, pp. 6640--6651.

\bibitem{schusteretal2021consistent}
T.~Schuster, A.~Fisch, T.~Jaakkola, and R.~Barzilay, ``Consistent accelerated
  inference via confident adaptive transformers,'' in \emph{EMNLP}, 2021, pp.
  4962--4979.

\bibitem{wei2022uncertainty}
L.~Wei, D.~Hu, W.~Zhou, and S.~Hu, ``Uncertainty-aware propagation structure
  reconstruction for fake news detection,'' in \emph{COLING}, 2022, pp.
  2759--2768.

\bibitem{chen2019embedding}
X.~Chen, M.~Chen, W.~Shi, Y.~Sun, and C.~Zaniolo, ``Embedding uncertain
  knowledge graphs,'' in \emph{AAAI}, 2019, pp. 3363--3370.

\bibitem{chen2021probabilistic}
X.~Chen, M.~Boratko, M.~Chen, S.~S. Dasgupta, X.~L. Li, and A.~McCallum,
  ``Probabilistic box embeddings for uncertain knowledge graph reasoning,'' in
  \emph{NAACL}, 2021, pp. 882--893.

\bibitem{boutouhami2020uncertain}
K.~Boutouhami, J.~Zhang, G.~Qi, and H.~Gao, ``Uncertain ontology-aware
  knowledge graph embeddings,'' in \emph{JIST}, 2020, pp. 129--136.

\bibitem{li2022rethinking}
Y.~Li, L.~Liu, and S.~Shi, ``Rethinking negative sampling for handling missing
  entity annotations,'' in \emph{ACL}, 2022, pp. 7188--7197.

\bibitem{wang2019improving}
S.~Wang, Y.~Liu, C.~Wang, H.~Luan, and M.~Sun, ``Improving back-translation
  with uncertainty-based confidence estimation,'' in \emph{EMNLP-IJCNLP}, 2019,
  pp. 791--802.

\bibitem{zhou2020uncertainty}
Y.~Zhou, B.~Yang, D.~F. Wong, Y.~Wan, and L.~S. Chao, ``Uncertainty-aware
  curriculum learning for neural machine translation,'' in \emph{ACL}, 2020,
  pp. 6934--6944.

\bibitem{zhang2021knowing}
S.~Zhang, C.~Gong, and E.~Choi, ``Knowing more about questions can help:
  Improving calibration in question answering,'' in \emph{ACL-IJCNLP}, 2021,
  pp. 1958--1970.

\bibitem{kertkeidkachorn2020gtranse}
N.~Kertkeidkachorn, X.~Liu, and R.~Ichise, ``Gtranse: generalizing
  translation-based model on uncertain knowledge graph embedding,'' in
  \emph{JSAI}, 2020, pp. 170--178.

\bibitem{glushkova2021uncertainty}
T.~Glushkova, C.~Zerva, R.~Rei, and A.~F. Martins, ``Uncertainty-aware machine
  translation evaluation,'' in \emph{EMNLP-Findings}, 2021, pp. 3920--3938.

\bibitem{jiang2021can}
Z.~Jiang, J.~Araki, H.~Ding, and G.~Neubig, ``How can we know when language
  models know? on the calibration of language models for question answering,''
  \emph{TACL}, pp. 962--977, 2021.

\bibitem{lin2022teaching}
S.~Lin, J.~Hilton, and O.~Evans, ``Teaching models to express their uncertainty
  in words,'' 2022.

\bibitem{zhou2023navigating}
K.~Zhou, D.~Jurafsky, and T.~Hashimoto, ``Navigating the grey area: Expressions
  of overconfidence and uncertainty in language models,'' \emph{arXiv preprint
  arXiv:2302.13439}, 2023.

\bibitem{shen2019modelling}
A.~Shen, D.~Beck, B.~Salehi, J.~Qi, and T.~Baldwin, ``Modelling uncertainty in
  collaborative document quality assessment,'' in \emph{Proceedings of the 5th
  Workshop on Noisy User-generated Text (W-NUT 2019)}, 2019, pp. 191--201.

\bibitem{jean2015montreal}
S.~Jean, O.~Firat, K.~Cho, R.~Memisevic, and Y.~Bengio, ``{M}ontreal neural
  machine translation systems for {WMT}{'}15,'' in \emph{Proceedings of the
  Tenth Workshop on Statistical Machine Translation}, 2015, pp. 134--140.

\bibitem{koehn2017sixChallenges}
P.~Koehn and R.~Knowles, ``Six challenges for neural machine translation,'' in
  \emph{Proceedings of the First Workshop on Neural Machine Translation}, 2017,
  pp. 28--39.

\bibitem{wang2020inference}
S.~Wang, Z.~Tu, S.~Shi, and Y.~Liu, ``On the inference calibration of neural
  machine translation,'' in \emph{ACL}, 2020, pp. 3070--3079.

\bibitem{newberry2021parliamentary}
T.~Newberry and T.~Ord, ``The parliamentary approach to moral uncertainty.''

\bibitem{hendrycks2023natural}
D.~Hendrycks, ``Natural selection favors ais over humans,'' 2023.

\end{thebibliography}
\end{document}